\documentclass[sansbold=false]{phai}
\usepackage{tabularx}
\usepackage{longtable}
\usepackage{needspace}
\usepackage{placeins}
\usepackage{pdflscape}
\usepackage{algorithm}
\usepackage{algpseudocode}
\usepackage{listings}
\usepackage{adjustbox}
\newtheorem{proposition}{Proposition}
\newtheorem{definition}{Definition}
\crefname{proposition}{proposition}{propositions}
\Crefname{proposition}{Proposition}{Propositions}
\crefname{definition}{definition}{definitions}
\Crefname{definition}{Definition}{Definitions}
\newcommand{\E}{\mathbb{E}}
\newcolumntype{Y}{>{\raggedright\arraybackslash}X}

\title{DiscoveryRSI: Recursively Learning to Discover Valuable Hypotheses}
\subtitle{}
\authors{\fontsize{11}{12}\selectfont
\adjustbox{max width=\linewidth}{%
\begin{tabular}{@{}c@{}}
Jianyuan Zhong\textsuperscript{1,2,*}\enspace
Chenwei Wu\textsuperscript{3,*}\enspace
Zhaochen Yu\textsuperscript{4,*}\enspace
Xixian Jiang\textsuperscript{11}\enspace
Jiajun Fan\textsuperscript{14}\enspace
Xuewei Li\textsuperscript{6}\enspace
Hao Xiao\textsuperscript{5}\\
Xilin Dai\textsuperscript{2}\enspace
Junyang Wang\textsuperscript{7}\enspace
Jiaru Zou\textsuperscript{8}\enspace
Zhiqing Cui\textsuperscript{9}\enspace
Boyuan Zheng\textsuperscript{3}\enspace
Jing Xiong\textsuperscript{10}\enspace
Zhen Zhang\textsuperscript{10}\\
Qiang Gao\textsuperscript{5}\enspace
Michael Morley\textsuperscript{12}\enspace
Luis Nakayama\textsuperscript{13,14}\enspace
Qiang Xu\textsuperscript{2}\enspace
Liyue Shen\textsuperscript{3}\enspace
Tao Gui\textsuperscript{5}\\
Leo Anthony Celi\textsuperscript{12,14,\textdagger}\enspace
Zhenfei Yin\textsuperscript{1,15,\textdagger}\enspace
Yingcheng Wu\textsuperscript{1,8,\textdagger}\enspace
Ling Yang\textsuperscript{1,7,16,\textdagger}
\end{tabular}%
}}

\affiliations{
\textsuperscript{1}PhAI Labs\quad
\textsuperscript{2}The Chinese University of Hong Kong\quad
\textsuperscript{3}University of Michigan\\
\textsuperscript{4}National University of Singapore\quad
\textsuperscript{5}Fudan University\quad
\textsuperscript{6}Peking University People's Hospital\\
\textsuperscript{7}Peking University\quad
\textsuperscript{8}Stanford University\quad
\textsuperscript{9}HKUST\quad
\textsuperscript{10}University of California, Santa Barbara\\
\textsuperscript{11}Southeast University\quad
\textsuperscript{12}Harvard Medical School\quad
\textsuperscript{13}Federal University of S\~ao Paulo
\textsuperscript{14}UIUC\\
\textsuperscript{15}Massachusetts Institute of Technology\quad
\textsuperscript{16}University of Oxford\quad
\textsuperscript{17}Princeton University}

\legend{\textsuperscript{*}Equal contribution\quad
\textsuperscript{\textdagger}Corresponding authors}
\runningtitle{DiscoveryRSI: Recursively Learning to Discover Valuable Hypotheses}
\hypersetup{pdftitle={DiscoveryRSI},pdfauthor={Jianyuan Zhong, Chenwei Wu, Zhaochen Yu, Xixian Jiang, Jiajun Fan, Hao Xiao, Xuewei Li, Xilin Dai, Junyang Wang, Jiaru Zou, Zhiqing Cui, Boyuan Zheng, Jing Xiong, Zhen Zhang, Qiang Gao, Michael Morley, Luis Nakayama, Qiang Xu, Liyue Shen, Tao Gui, Leo Anthony Celi, Zhenfei Yin, Yingcheng Wu, Ling Yang},pdfsubject={Agentic hypothesis discovery, trajectory learning, and DFM-Bench}}
\makeatletter
\renewcommand\phai@titlefont{\linespread{1}\fontsize{20}{23}\selectfont}
\makeatother
\DeclareCaptionFont{teasercaption}{\fontsize{9}{10.5}\selectfont}
\titlespacing*{\section}{0pt}{1.25\baselineskip}{0.35\baselineskip}
\titlespacing*{\subsection}{0pt}{0.90\baselineskip}{0.25\baselineskip}
\titlespacing*{\paragraph}{0pt}{0.60\baselineskip}{0.5em}
\selectfont
\begin{document}
\maketitle
\vspace{-20pt}
\enlargethispage{10mm}
\begin{center}
{\sffamily\bfseries\fontsize{10}{12}\selectfont\color{phaiaccent}
Code: \href{https://github.com/Gen-Verse/DiscoveryRSI}{GitHub} \quad\textbar\quad Models: \href{https://huggingface.co/collections/Gen-Verse/discoveryrsi}{Hugging Face} \quad\textbar\quad Website: \href{https://foundry-arena.rsi-arena.workers.dev/arena}{DiscoveryRSI Arena}\par}
\end{center}
\vspace{-3mm}
\begingroup
\captionsetup{font=teasercaption}
\teaser{%
  \includegraphics[width=\linewidth]{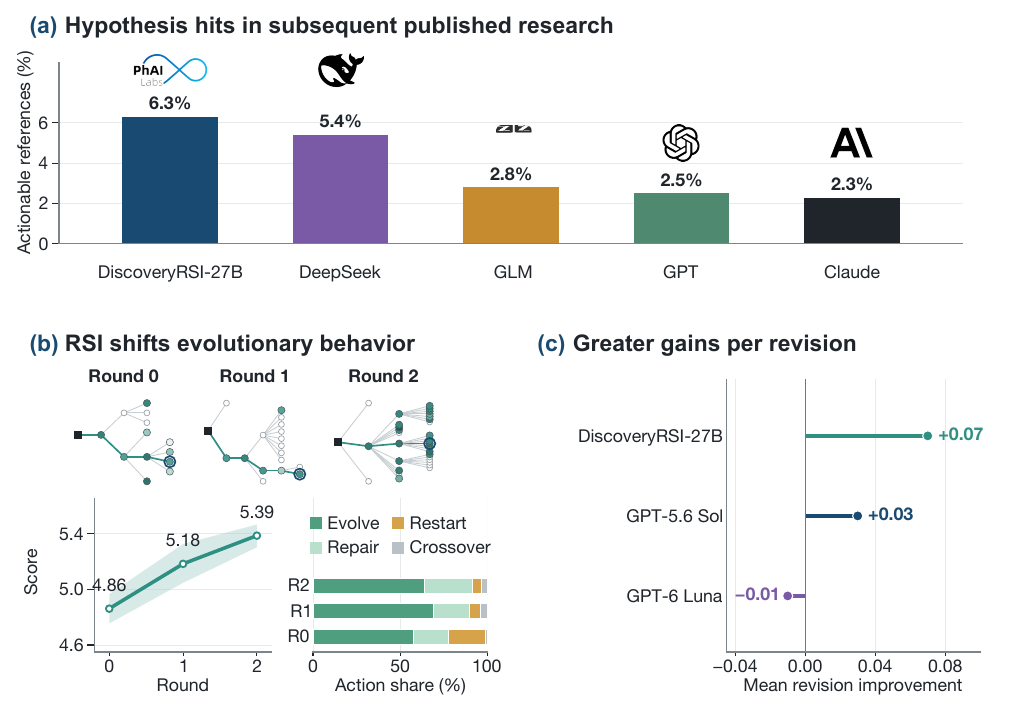}%
}{\textbf{(a) More actionable links to published research.} The rate is the fraction of generated hypotheses linked to post-cutoff expert publications supplying a concrete design change, comparator, measurement, or testable evidence gap. DiscoveryRSI leads at 6.3\% (\Cref{app:reference-actionability}). \textbf{(b) RSI improves quality and shifts evolutionary behavior.} Score measures research quality and action share is the fraction of episodes using each operation. Research quality rises from 4.86 to 5.39 as evolve/repair use rises from 77.5\% to 91.7\%. \textbf{(c) Larger human-rated gains per revision.} Mean revision change averages the human-score difference between a hypothesis and its predecessor: $+0.07$ for DiscoveryRSI, $+0.03$ for Sol, and $-0.01$ for Luna. 
}
\endgroup
\clearpage
\begin{abstract}


Scientific discovery requires developing worthwhile hypotheses into practical investigations, yet learning from this evolving research process remains challenging. 
We introduce an operational definition of \textbf{valuable and testable research hypotheses}: evidence-grounded claims about consequential uncertainties, coupled with feasible investigations, explicit decision criteria, and alternative explanations.
Our \textbf{DiscoveryRSI} framework develops hypothesis trees through multi-agent evolutionary search, compiles selected investigations into executable Harbor tasks, and recursively learns through three mechanisms: \textbf{hypothesis credit optimization} attributes scientific assessments to documented research contributions; \textbf{harness replay} identifies productive strategies and recurring failures to refine research workflows; and \textbf{rediscovery} applies accepted improvements to new investigations, generating further experience for subsequent learning. We also introduce DFM-Bench, which provides comprehensive assessment of hypothesis quality and cumulative research progress at the levels of hypotheses, revisions, development paths and research runs. Across four life-science domains, automated evaluation shows that hypothesis quality rises from 4.86 for the base model to 5.39 after two rounds of self-improvement, with the large gains in falsifiability and data resolvability. We further validate hypothesis quality through expert review, comparing our \textbf{DiscoveryRSI-27B} with other research agents. Preliminary human assessments show that DiscoveryRSI-27B scores higher than GPT-6 Luna on hypothesis importance and whether claims are plausible and evidence-bounded. 
DiscoveryRSI-27B achieved a 6.3\% rate of hypothesis correspondence with research published after the model cutoff, largely outperform GPT-6 with 2.5\% and Claude Opus 5 with 2.3\%, while delivering greater human-rated improvement along research paths.
\end{abstract}
\section{Introduction}
Research requires developing worthwhile hypotheses into practical investigations. Developing such hypotheses involves gathering evidence, revising explanations, and deciding which directions merit further effort. This process generates experience about both the hypotheses under investigation and the decisions that guide their development, which allows human scientists to better design their next study. For research agents, using this experience to improve future research requires identifying useful contributions and translating them into changes to the model and research workflow. This motivates our central research question: \textit{How can research agents turn hypothesis development and investigation into a recursive learning process that improves subsequent research?}

\begin{figure}[p]
\centering
\makebox[\linewidth][c]{%
\begin{minipage}{170mm}
\centering
\includegraphics[width=\linewidth,height=0.74\textheight,keepaspectratio]{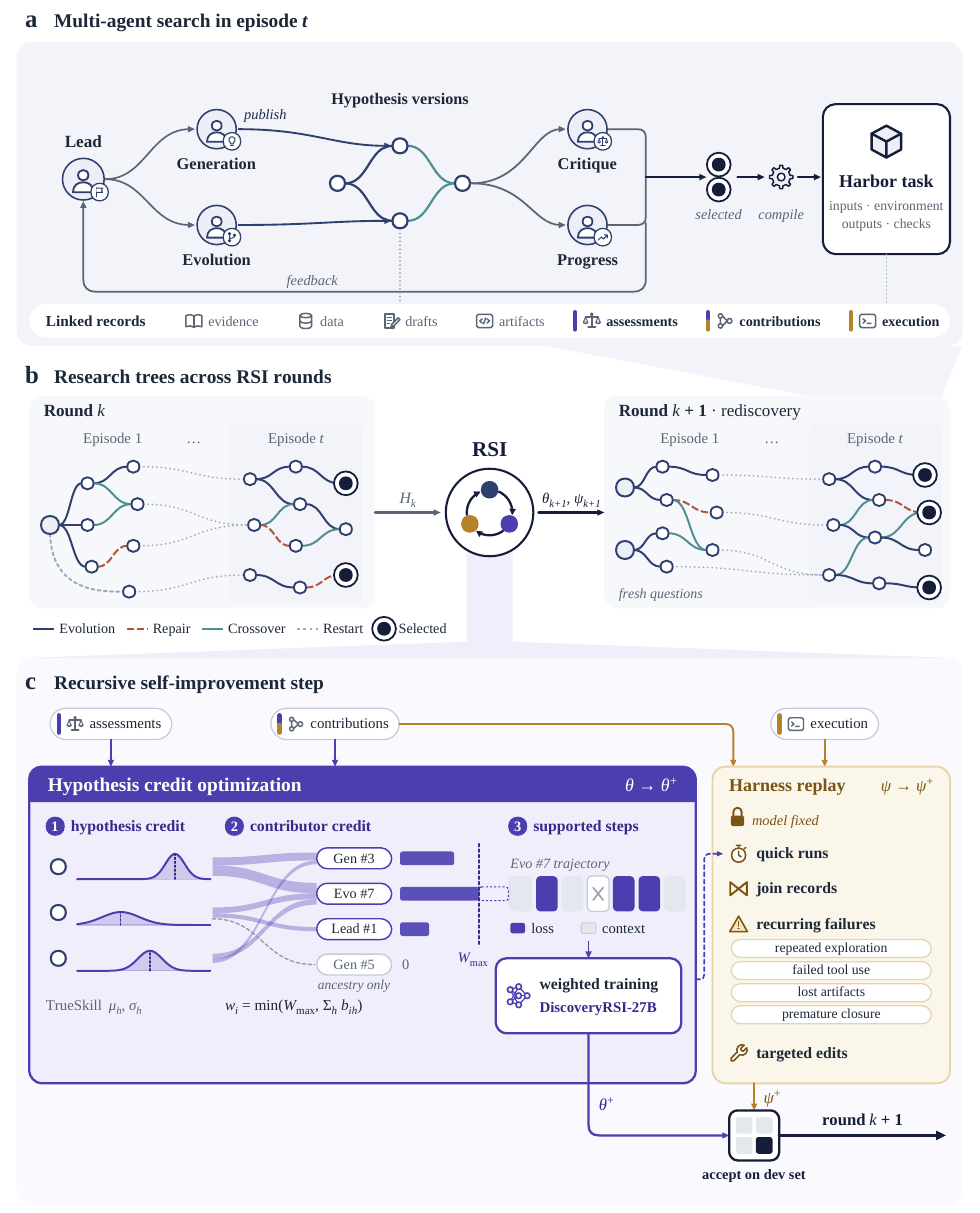}
\caption{DiscoveryRSI couples multi-agent hypothesis development with recursive self-improvement. \textbf{(a)} Within an episode, a Lead coordinates Generation and Evolution; candidate hypothesis versions are critiqued and progress is recorded. Selected hypotheses are compiled into Harbor tasks, with linked evidence, data, drafts, artifacts, assessments, contributions, and execution records. \textbf{(b)} Research trees carry hypotheses across RSI rounds: the recorded state $H_k$ enters RSI, which produces updated parameters $(\theta_{k+1},\psi_{k+1})$ and fresh questions for rediscovery. \textbf{(c)} Two update branches learn from the same experience: \emph{hypothesis credit optimization} turns assessments and documented contributions into bounded weighted training, while \emph{harness replay} keeps the model fixed, diagnoses recurring failures, and tests targeted workflow edits. Accepted updates are checked on the development set before round $k+1$.}
\label{fig:discovery-system}
\end{minipage}%
}
\end{figure}
Addressing this question requires connecting hypothesis development and experimental execution with learning that improves the agent over investigations. Co-Scientist \citep{coscientist} develops hypotheses through multi-agent evolutionary refinement, while AI Scientist-v2 \citep{aiscientistv2} connects hypothesis generation with experimental execution through agentic tree search. ScienceIDE \citep{scienceide} constructs learning environments from scientific codebases, and ScienceBuddy \citep{sciencebuddy} turns researcher interactions into executable tasks and feedback for model and harness improvement. Building on these approaches, we focus on learning from hypotheses that agents actively develop: preserving how proposals evolve, linking assessments of research quality to the contributions behind them, and using this experience to improve subsequent investigations.

Turning hypothesis development into a recursive learning process presents challenges in both execution and learning. Translating a proposal into an executable investigation requires resolving its data requirements and verification criteria while preserving the research question. Learning from the resulting research experience involves a further challenge: a hypothesis's final assessment does not specify which contributions should be reinforced. One agent may propose an explanation, another introduce a useful control, and a third revise the investigation plan in response to new evidence. Their contributions can remain obscured if only the final proposal is retained. Addressing these challenges involves making investigation plans concrete while preserving the development histories that connect assessments to the actions behind them.

To address these challenges, we introduce \textbf{DiscoveryRSI}, a general framework that turns hypothesis-driven research into a recursive learning process (\Cref{fig:discovery-system}). It develops research trees of testable hypotheses through multi-agent evolutionary search, preserving the evidence and decisions behind each proposal and revision. Selected computational investigations are compiled into executable Harbor tasks with concrete inputs, environments, required outputs, and verification checks. The retained research histories support three coupled recursive improvement mechanisms. \textbf{Hypothesis credit optimization} links assessments of research quality to documented contributions and assigns bounded credit to the contributing histories, providing supervision for the associated model actions. We use this supervision to train \textbf{DiscoveryRSI-27B}. \textbf{Harness replay} examines these histories to identify productive strategies and recurring failures, informing changes to agent coordination, evidence access, and research allocation. \textbf{Rediscovery} applies accepted model and harness updates to fresh investigations, generating new hypotheses and experience for subsequent learning.

We introduce \textbf{DFM-Bench} to evaluate hypothesis quality and cumulative research progress. It assesses whether proposals are worthwhile and testable, what revisions contribute, and how development paths and research runs accumulate evidence and explore different directions. These assessments use hypotheses and their supporting research records, with the execution readiness of compiled Harbor tasks measured separately.

Our contributions are fourfold:
\begin{itemize}
\item \textbf{An operational definition and a general framework for hypothesis-driven research.} We define valuable and testable research hypotheses as joint objects comprising a claim, an investigation plan, decision criteria, and alternative explanations (\Cref{def:valuable-testable}). DiscoveryRSI develops these objects through multi-agent evolutionary search, compiles selected investigations into executable Harbor tasks, and preserves the underlying research decisions and development histories as reusable learning experience.
\item \textbf{Recursive self-improvement from research experience.} We develop hypothesis credit optimization to attribute assessments of research quality to documented contributions and train DiscoveryRSI-27B, harness replay to refine research strategies and workflows, and rediscovery to apply accepted improvements to subsequent investigations.
\item \textbf{A benchmark for hypothesis quality and cumulative research progress.} DFM-Bench evaluates hypotheses, revisions, development paths, research runs, and research contributions. We further evaluate DiscoveryRSI through expert-supervised review and controlled model comparisons with its base model and other research agents. Results show clear gains over rounds of self-improvements and changes in agent behavior patterns.
\item \textbf{Empirical evidence for learning from scientific research.} DiscoveryRSI increase mean hypothesis quality from 4.86 to 5.39, with gains in falsifiability, alternative explanations, and data resolvability. Expert-supervised assessment shows greater human-rated improvement along research paths, with an average gain per revision of $+0.07$, compared with $+0.03$ for GPT-5.6 Sol and $-0.01$ for GPT-6 Luna. DiscoveryRSI-27B also achieves a 6.3\% rate of hypothesis correspondence with post-cutoff research, compared with 2.5\% for GPT-6 and 2.3\% for Claude Opus 5. Together, these findings show that learning from research histories improves the formulation of testable investigations and the development of hypotheses toward concrete research opportunities.
\end{itemize}

\Needspace{8\baselineskip}
\section{Valuable and Testable Research Hypotheses}
\label{sec:hypotheses}
\label{sec:setting}

\subsection{An Operational Definition}

We introduce a joint operational definition of \textbf{valuable and testable research hypotheses} for hypothesis-driven research agents. It makes the scientific value of a claim and the evidential capacity of its investigation explicit within one revisable research object.

\begin{definition}[Valuable and testable research hypothesis]
\label{def:valuable-testable}
\normalfont
Relative to a documented evidence base $\mathcal E$ and declared data, experimental, and resource constraints $\mathcal B$, a research hypothesis is represented as
\begin{equation}
\mathcal H=(c,\pi,d,a),
\label{eq:hypothesis-definition}
\end{equation}
where $c$ is an evidence-grounded claim, $\pi$ an investigation plan, $d$ its decision criteria, and $a$ the relevant alternative explanations. It is \textbf{valuable} when resolving the uncertainty in $c$ could make a consequential addition to current understanding or change a justified research decision relative to $\mathcal E$. It is \textbf{testable} when $\pi$ specifies an investigation feasible under $\mathcal B$ whose observable outcomes, interpreted through $d$ and $a$, can provide evidence that supports, weakens, or bounds $c$, with unresolved distinctions stated explicitly. A valuable and testable research hypothesis satisfies both requirements.
\end{definition}

The four components make this definition operational. The \textbf{claim} specifies the system or population, scope, observable quantities, and scientific uncertainty. The \textbf{investigation plan} identifies the data or measurements, comparisons, controls, analysis, and required dependencies. The \textbf{decision criteria} state, before interpreting results, which outcomes would support or weaken the claim and which would remain inconclusive. The \textbf{alternative explanations} identify competing accounts, confounding, or measurement limitations, together with the comparisons that can distinguish them and the limits of those comparisons.

Scientific value can arise from a new explanation, a useful replication, a measurement check, or an informative negative result when it changes understanding or research priorities. Testability depends on the declared evidence and constraints: unavailable measurements or experiments remain explicit dependencies until they can be obtained. These requirements assess a proposed research opportunity; they do not establish that the claim is true. DFM-Bench evaluates the resulting objects through its scientific-quality profile (\Cref{sec:dfm}), while completed execution and independent support are recorded separately.

\subsection{From Observations to Research Questions}

DiscoveryRSI focuses on how a research question is formed and tied to available evidence. IdeaScientist develops proposals from a specified problem and challenge, including evaluation plans and falsifiable predictions \citep{ideascientist2026}. Our emphasis is the joint development of a hypothesis and the investigation that can examine it: observations and data constraints shape the question, its comparisons, alternative explanations, and interpretation.

Descriptive analyses characterize observed patterns, variation, and measurement coverage, helping identify questions worth pursuing and the evidence available to investigate them. Each proposal then ties its claim to a defined system or population, observable quantities, and interpretable outcomes.

\subsection{A Hypothesis with an Investigation Plan}

\begin{figure}[!htbp]
\centering
\includegraphics[width=\linewidth]{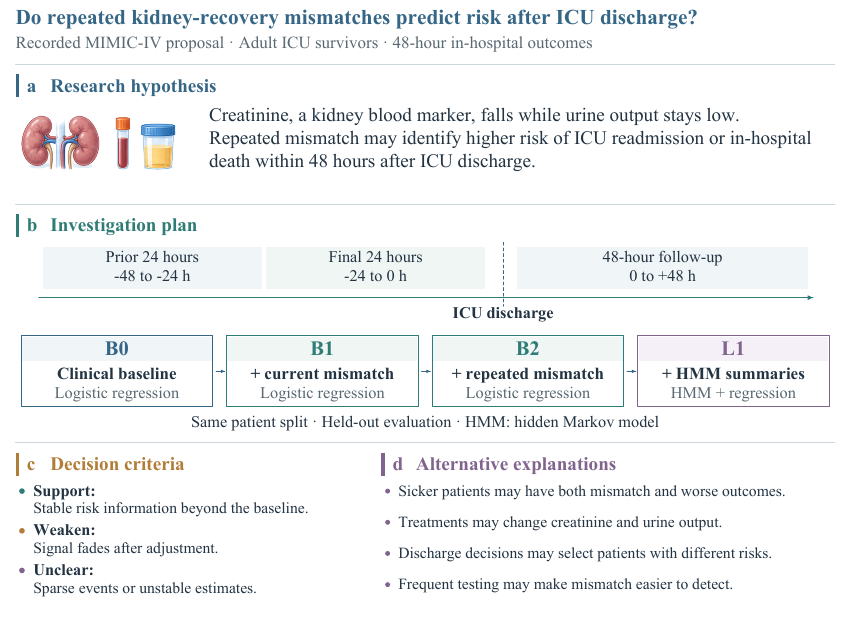}
\caption{\textbf{A recorded research hypothesis and its planned investigation.} This MIMIC-IV proposal concerns adults surviving a first ICU stay of at least 48 hours. \textbf{(a)} It asks whether repeated creatinine--urine output mismatch adds information about ICU readmission or in-hospital death within 48 hours after ICU discharge, during the same hospitalization. \textbf{(b)} B0--B2 compare clinical, current-mismatch, and repeated-mismatch logistic models; L1 adds hidden Markov model summaries, subject to data-support and stability checks. Models share a patient-level training, tuning, and test split. \textbf{(c,d)} Decision criteria and alternatives bound interpretation. No clinical outcome is asserted.}
\label{fig:hypothesis-formulation}
\end{figure}

\Cref{fig:hypothesis-formulation} instantiates \Cref{def:valuable-testable} with a MIMIC-IV proposal. Its value lies in asking whether repeated creatinine--urine output mismatch adds information about adverse outcomes after ICU discharge. Its testability comes from a defined population and outcome window, patient-level held-out comparisons against simpler baselines, explicit interpretation criteria, and alternative explanations that bound the claim. A supported lack of added information would also inform whether more complex models merit further investigation.

Recorded hypothesis versions retain links to their predecessors and supporting evidence. These records make changes in the question, investigation plan, and interpretation traceable, providing the research histories used by the trees and learning mechanisms in the following section.
\section{DiscoveryRSI}
\label{sec:system}
\label{sec:d2d}
DiscoveryRSI develops research hypotheses across episodes and retains the work behind each version as linked research records (\Cref{fig:discovery-system}). These records connect scientific assessments to contributing actions for model training, and connect execution failures to changes in the research harness. Selected investigations become executable Harbor tasks. Hypothesis credit optimization, harness replay, and rediscovery then turn this accumulated experience into a recursive learning process. Algorithm~\ref{alg:dsi-loop} summarizes the complete protocol, from within-run hypothesis development to accepted updates and fresh collection.

\subsection{Research Trees and Hypothesis Development}

\paragraph{Research hypothesis trees.}
Each node is a versioned hypothesis with its investigation plan, decision criteria, alternative explanations, and evidence. Parent links record scientific development: a revision has a predecessor, while a combination can have several parents. The ancestry can therefore form a directed acyclic graph, although we use research tree as an intuitive description. It spans episodes and remains distinct from the sequence of agent actions that produced each node.

\paragraph{Linked research records.}
The hypothesis tree is part of a broader network linking hypotheses to sources and data, drafts and supporting artifacts, scientific assessments, agent contributions, and execution records. Files preserve the research material; registry entries, identifiers, paths, and available content hashes preserve its relationships. A shared artifact can inform several hypotheses without making them scientific descendants of one another. These links let us trace an assessed contribution to the histories that produced it, or trace a recurring failure to the decisions and execution events behind it. Appendix~\ref{app:linked-records} specifies the record types and link semantics.

\paragraph{Growing through evolutionary operations.}
Following operator-based search such as Frontis-MA1 \citep{frontisma1}, evolution develops a hypothesis, repair addresses an identified defect, and crossover combines contributions from multiple parents. Scientific restart opens a new question without assigning parents; earlier work may remain as context. Lead selects directions, parents, and operations. Generation and Evolution investigate evidence and publish proposals or revisions. Blinded Critique compares scientific merit, while ancestor-aware Progress assesses substantive progress, necessary repair, repetition, or insufficient evidence. Their assessments guide subsequent work and are retained in the linked records.

\subsection{From Hypotheses to Executable Investigations}

Selected computational plans are compiled into Harbor tasks \citep{harbor}, with concrete inputs, environments, required outputs, and verification checks. Compilation preserves the hypothesis's scientific commitments: the population or biological system, proposed comparison, discriminating outcomes, and interpretation criteria. Unavailable data and physical experiments remain explicit dependencies.

Execution checks establish whether a task satisfies its computational contract; scientific assessment determines what the investigation can establish. The compiled package remains linked to its selected hypothesis, inputs, artifacts, and validation records. Hypothesis development can therefore supply learning experience before a complete investigation has been executed.
\subsection{Hypothesis Credit Optimization}
\label{sec:dsi-distillation}

Hypothesis credit optimization follows three links: scientific assessments determine hypothesis credit; contribution records allocate it to documented research histories; supported actions in those histories become training targets (\Cref{fig:discovery-system}c).

\paragraph{Assessing hypotheses.}
Blinded Critique provides preferences, ties, or abstentions under compatible data and review conditions. TrueSkill \citep{trueskill} summarizes relative preference strength and uncertainty:
\begin{equation}
\begin{aligned}
z_h &\sim \mathcal N(\mu_{h,0},\sigma_{h,0}^{2}),
& t_{h,e}\mid z_h &\sim \mathcal N(z_h,\beta^2),\\
P(a\succ b\mid z_a,z_b)
&=\Phi\!\left(
\frac{z_a-z_b-\varepsilon}{\sqrt{2}\beta}
\right),\\
p(z_h\mid\mathcal D_{\mathrm{cmp}})
&\approx \mathcal N(\mu_h,\sigma_h^2).
\end{aligned}
\label{eq:trueskill-summary}
\end{equation}
Here $z_h$ is rubric-conditioned preference strength, $t_{h,e}$ its noisy comparison performance, $\beta$ the noise scale, $\varepsilon$ the tie margin, and $\Phi$ the standard normal CDF. Insufficient evidence supplies no preference update. The resulting $(\mu_h,\sigma_h)$ describe comparative assessment, not the probability that a hypothesis is true. Separate Progress judgments assess changes relative to actual predecessors (Appendix~\ref{app:trueskill}).

\paragraph{Allocating contributor credit.}
Assessment coverage, uncertainty, validity, and recorded progress determine a bounded hypothesis weight. The linked contribution records then identify histories that produced a proposal, supplied a supporting investigation, or made a documented coordination decision. One hypothesis can receive contributions from several Generation or Evolution histories and from Lead assignments. For history $i$, the accumulated credit is
\begin{equation}
w_i=\min\!\left(W_{\max},\sum_h b_{ih}\right),
\label{eq:trajectory-credit}
\end{equation}
where $b_{ih}$ is the allocation from hypothesis $h$ after contributor sharing, duplicate suppression, and scientific-family budget limits. A history is a particular recorded invocation, not all work by a role. Unsupported shares remain unallocated; ancestry or participation alone earns no credit. Appendices~\ref{app:credit} and~\ref{app:record-attribution} give the allocation rules and provenance requirements.

\paragraph{Reinforcing supported actions.}
Credit controls how strongly a history contributes to training. Within it, the target set selects the assistant responses and tool-call arguments supported by the recorded contribution. This can include several critical steps across different role histories. Tool observations and known failed actions remain available as context but receive no direct loss. We train DiscoveryRSI-27B with
\begin{equation}
\mathcal L(\theta)=-
\sum_i\frac{\omega_i}{|U_i|}
\sum_{u\in U_i}
\log\pi_\theta(a_{iu}\mid x_{iu}),
\label{eq:sft}
\end{equation}
where $U_i$ contains eligible assistant-token positions, $x_{iu}$ is their recorded context under the serialization policy in \Cref{app:training}, and $\omega_i$ combines contributor credit with domain and shared-memory-group balancing. Thus, the history's weight is distributed over its supported targets, rather than over every message. Histories without eligible targets are excluded. Allocation and serialization details appear in Appendices~\ref{app:credit} and~\ref{app:training}.

\subsection{Harness Replay}
\label{sec:harness-adaptation}

Harness replay improves the research workflow while holding model weights fixed. ReAct formalizes interleaved reasoning and acting \citep{react}; Self-Harness and Agentic Harness Engineering study improvement from deliberately minimal tool-using configurations \citep{selfharness,agenticharness}. Our agents start from mature Codex or Claude Code harnesses. Their session, context, filesystem, and tool-execution mechanisms provide the substrate for complex investigations that continue across many actions. We adapt research coordination, evidence access, review timing, and work allocation on top of that substrate (Appendix~\ref{app:harness-substrate}).

Short diagnostic runs, including runs of a newly trained model candidate, first reveal where work breaks down. Agent contribution and decision records are joined with execution and delivery records to locate repeated exploration, failed tool use, lost artifacts, or premature closure. These diagnoses motivate targeted harness edits. For histories $\mathcal H_k$ collected with fixed model $\theta_k$ and harness $\psi_k$, candidate selection is
\begin{equation}
\begin{aligned}
\Psi_k^{\mathrm{cand}}
&=\operatorname{Propose}(\psi_k;\mathcal H_k),\\
\psi_k^{+}
&=\operatorname{Select}_{\mathrm{dev}}
\!\left(
\{\psi_k\}\cup\Psi_k^{\mathrm{cand}}
\,\middle|\,\theta_k,B
\right),
\end{aligned}
\label{eq:harness-replay}
\end{equation}
where $B$ denotes matched resource limits. Quick runs and supported replay diagnose problems and screen changes; matched development evaluation determines acceptance, considering scientific quality, exploration breadth, reliability, and cost. The incumbent remains eligible. Replay uses recorded observations and continuations; an unrecorded alternative requires execution from a compatible restored state. Appendices~\ref{app:harness-diagnosis} distinguish diagnosis, replay, and controlled comparison.

\subsection{Rediscovery}
\label{sec:recursive-redeployment}

Rediscovery applies accepted model and harness updates to fresh investigations. Let $\theta_k^{+}$ be the model candidate from hypothesis credit optimization and $\psi_k^{+}$ the harness candidate selected through replay and development evaluation. The next cycle is
\begin{equation}
\begin{aligned}
(\theta_{k+1},\psi_{k+1})
&=\operatorname{Accept}_{\mathrm{dev}}
\!\left(
\{\theta_k,\theta_k^{+}\}
\times
\{\psi_k,\psi_k^{+}\}
\right),\\
\mathcal H_{k+1}
&\sim\operatorname{Research}
\!\left(
\theta_{k+1},\psi_{k+1};
\mathcal Q_{k+1}^{\mathrm{fresh}},B
\right).
\end{aligned}
\label{eq:rediscovery}
\end{equation}
The acceptance procedure evaluates qualifying updates separately and jointly, so the model, harness, both, or neither may change. The accepted system returns to fresh questions and collects new hypotheses, evidence, assessments, and contribution histories. Scientific restart changes direction within a research run; rediscovery changes the system used for the next run. Versioned records connect successive cycles while keeping development and final evaluation families separate (Appendices~\ref{app:harness-diagnosis}).

\begin{algorithm}[!htbp]
\caption{DiscoveryRSI: recursive improvement from hypothesis development}
\label{alg:dsi-loop}
\footnotesize
\begin{algorithmic}[1]
\Require $(\theta_0,\psi_0)$, $\{\mathcal Q_k^{\mathrm{fresh}}\}_{k=0}^{K-1}$, $\mathcal Q_{\mathrm{dev}}$, $\mathcal Q_{\mathrm{test}}$, budget $B$
\For{$k=0,\ldots,K-1$}
  \State $(G_k,\mathcal H_k)\sim\operatorname{Research}(\theta_k,\psi_k;\mathcal Q_k^{\mathrm{fresh}},B)$ \Comment{Tree and histories}
  \State $\mathcal T_k\gets\operatorname{CompileHarbor}(\operatorname{Select}(G_k))$
  \State $\{(\mu_h,\sigma_h,\Delta_h)\}_{h\in G_k}\gets\operatorname{Assess}(G_k,\mathcal H_k)$ \Comment{Merit and progress}
  \State $\{b_{ih}\}\gets\operatorname{Credit}(\{(\mu_h,\sigma_h,\Delta_h)\},\mathcal H_k)$ \Comment{Bounded attribution; Appendix~\ref{app:credit}}
  \State $w_i\gets\min\!\left(W_{\max},\sum_h b_{ih}\right)$
  \State $U_i\gets\operatorname{SupportedTargets}(i,\mathcal H_k)$; $\{\omega_i\}\gets\operatorname{Balance}(\{w_i:|U_i|>0\})$
  \State $\displaystyle\mathcal L_k(\theta)=-\sum_{i:\,|U_i|>0}\frac{\omega_i}{|U_i|}\sum_{u\in U_i}\log\pi_\theta(a_{iu}\mid x_{iu})$ \Comment{Eq.~\eqref{eq:sft}}
  \State $\theta_k^+\gets\operatorname{Update}(\theta_k;\mathcal L_k)$ \Comment{$\psi_k$ fixed}
  \State $\Psi_k^{\mathrm{cand}}\gets\operatorname{Propose}(\psi_k;\mathcal H_k)$ \Comment{Supported replay}
  \State $\psi_k^+\gets\operatorname{Select}_{\mathrm{dev}}(\{\psi_k\}\cup\Psi_k^{\mathrm{cand}}\mid\theta_k,\mathcal Q_{\mathrm{dev}},B)$ \Comment{$\theta_k$ fixed}
  \State $(\theta_{k+1},\psi_{k+1})\gets\operatorname{Accept}_{\mathrm{dev}}\!\left(\{\theta_k,\theta_k^+\}\times\{\psi_k,\psi_k^+\}\mid\mathcal Q_{\mathrm{dev}},B\right)$
\EndFor
\State $S_{\mathrm{test}}\gets\operatorname{Eval}(\theta_K,\psi_K;\mathcal Q_{\mathrm{test}},B)$ \Comment{Untouched test families}
\State \Return $(\theta_K,\psi_K),\ S_{\mathrm{test}}$
\end{algorithmic}
\end{algorithm}

\section{DFM-Bench: Evaluating Scientific Quality and Research Progress}
\label{sec:dfm}

DFM-Bench evaluates scientific inquiry through five connected questions: whether a hypothesis is worthwhile and testable, what a revision contributes, whether a development path accumulates useful evidence, how a research run allocates effort, and what distinct scientific contribution emerges. Each question requires evidence appropriate to its level.

\begin{table}[!htbp]
\centering\small
\caption{\textbf{Evaluation coverage of selected scientific-discovery benchmarks.} A check denotes explicit assessment, a circle related or partial assessment, and a cross no explicit assessment under our definitions. Recording intermediate actions alone does not establish coverage. Definitions and qualifications appear in Appendix~\ref{app:dfm-comparison}.}
\label{tab:benchmark-coverage}
\begingroup
\setlength{\tabcolsep}{3pt}
\newcommand{\covered}{\textcolor{Teal}{\ensuremath{\checkmark}}}
\newcommand{\partialcoverage}{\textcolor{DarkGoldenrod}{\ensuremath{\circ}}}
\newcommand{\nocoverage}{\textcolor{red}{\ensuremath{\times}}}
\begin{tabularx}{\linewidth}{@{}p{3.3cm}*{6}{>{\centering\arraybackslash}X}@{}}
\toprule
& \multicolumn{2}{c}{Scientific assessment} & \multicolumn{4}{c}{Research-process assessment} \\
\cmidrule(lr){2-3}\cmidrule(lr){4-7}
Benchmark & {\scriptsize Hypothesis merit} & {\scriptsize Contribution maturity} & {\scriptsize Evidence use} & {\scriptsize Revision} & {\scriptsize Development path} & {\scriptsize Whole-run allocation} \\\midrule
AgentIdeaBench \citep{agentideabench} & \covered & \nocoverage & \partialcoverage & \nocoverage & \nocoverage & \nocoverage \\
DiscoveryBench \citep{discoverybench} & \partialcoverage & \nocoverage & \partialcoverage & \nocoverage & \nocoverage & \nocoverage \\
BLADE \citep{blade} & \nocoverage & \nocoverage & \partialcoverage & \nocoverage & \nocoverage & \nocoverage \\
ScienceAgentBench \citep{scienceagentbench} & \nocoverage & \nocoverage & \partialcoverage & \nocoverage & \nocoverage & \nocoverage \\
OEB \citep{oeb} & \nocoverage & \nocoverage & \covered & \covered & \partialcoverage & \partialcoverage \\
\midrule
\textbf{DFM-Bench (ours)} & \covered & \covered & \covered & \covered & \covered & \covered \\
\bottomrule
\end{tabularx}
\endgroup
\end{table}

\subsection{Five Levels of Scientific Assessment}

\paragraph{Hypothesis quality.}
Is an individual hypothesis worthwhile and testable? Assessment examines the proposal, investigation plan, and supporting evidence. It produces a scientific-quality profile covering importance, novelty, grounding, falsifiability, data resolvability, feasibility, alternative explanations, and actionability, together with an aggregate score when the required assessments are complete.

\paragraph{Revision value.}
What did a revision contribute? Assessment compares a hypothesis with its actual predecessor and examines the evidence and actions supporting the change. It distinguishes substantive progress, necessary repair, repetition, and insufficient evidence. Changes in hypothesis scores are recorded separately from these judgments.

\paragraph{Development-path progress.}
Does a development path accumulate useful evidence? Assessment follows an ordered sequence of actual ancestry relationships, examining evidence acquisition, coherent revision, scientific continuity, and decisions to continue or stop. Agent action histories provide supporting evidence for these judgments, while the development path records how the hypotheses themselves changed.

\paragraph{Research-run quality.}
Does a research run explore and allocate effort productively? Assessment examines the full research graph, selection and stopping decisions, and available opportunity and resource records. It considers the breadth and depth of exploration, synthesis across branches, and allocation of the research budget. Run-level quality is assessed directly rather than obtained by averaging path scores.

\paragraph{Scientific contribution.}
What distinct scientific contribution emerges across related proposals? Assessment groups proposals by scientific question and records what was added beyond the starting evidence: a new distinction, an informative negative result, a supported finding, or a reduction in uncertainty. Contribution maturity distinguishes proposed ideas from contributions supported by completed investigations and independent checks.

\subsection{Coverage Relative to Existing Benchmarks}

Existing benchmarks examine complementary aspects of scientific work, including ideation, goal-directed discovery, analytical decisions, and executable tasks \citep{agentideabench,discoverybench,blade,scienceagentbench}. OEB further evaluates evidence use, experimentation, and research-process integrity \citep{oeb}. DFM-Bench makes hypotheses, revisions, development paths, research runs, and contribution families explicit assessment objects. Table~\ref{tab:benchmark-coverage} compares their evaluation coverage; the accompanying definitions and qualifications appear in Appendix~\ref{app:dfm-comparison}.

\section{Expert-Supervised Assessment of Scientific Hypotheses}
\label{sec:expert-assessment}

We conduct expert-supervised human assessment to examine the scientific value and design quality of generated hypotheses. The review considers whether a proposed investigation addresses a worthwhile question, fits the available data, distinguishes alternative explanations, and yields interpretable outcomes.

The assessment involved 27 medical doctors, medical students and bioscience PhD students with prior biomedical or clinical research backgrounds, under three senior supervisors. The review cohort contains 292 proposals from six dataset groups spanning the four life-science domains: HCC clinical cohort (52), MIMIC-IV (60), eICU (60), UK Biobank (55), GEO targets (35), and GEO mechanisms (30). Proposals were drawn from a fixed hypothesis-atlas snapshot and presented in randomized order.

\begin{figure}[!htbp]
\centering
\includegraphics[width=\linewidth]{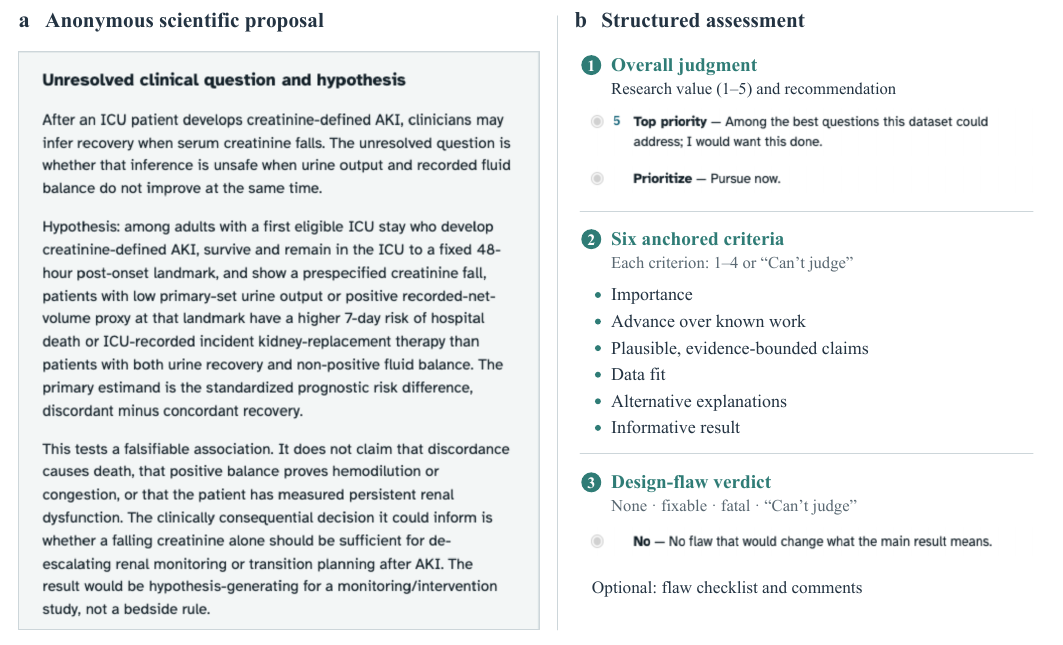}
\caption{\textbf{Expert-supervised assessment of scientific hypotheses.} \textbf{(a)} An anonymous MIMIC-IV proposal presents its scientific question, planned comparison, and claim limits. \textbf{(b)} Reviewers record an overall judgment, assess six anchored criteria, and identify design flaws; flaw details and comments are optional. The criterion list summarizes the rubric, while radio-option excerpts reproduce one saved assessment. Identifying metadata is omitted. The example illustrates the review procedure rather than an aggregate evaluation result.}
\label{fig:expert-review}
\end{figure}

\paragraph{Anonymous proposal packets.}
Reviewers received English proposal copies containing the scientific question, investigation plan, data requirements, alternative explanations, and supporting citations. Producer and execution provenance were withheld, while scientific methods and references were retained. AI reviewer grades were hidden, and reviewers were instructed to assess each proposal on its own merits without consulting its revision history in the atlas. The interface presented the proposal alongside the assessment form (\Cref{fig:expert-review}); \Cref{app:human-assessment} consolidates the review protocol.

\paragraph{Scientific judgments and design flaws.}
The protocol asks reviewers to record an overall research-value judgment first, using a five-point scale from not worth pursuing to top priority, together with a recommendation to prioritize, revise, defer, or reject the proposal. Reviewers then assess six criteria on anchored four-point scales: importance, advance over known work, plausible and evidence-bounded claims, data fit, alternative explanations, and informative results. A separate verdict identifies whether a consequential design flaw is absent, fixable, or fatal. ``Can't judge'' is available when evidence or relevant expertise is insufficient. A flaw checklist, a description of the main flaw, and further comments are optional.

\section{Experimental Setup}
\label{sec:experimental-setup}

\begin{figure}[!htbp]
\centering
\includegraphics[width=\linewidth]{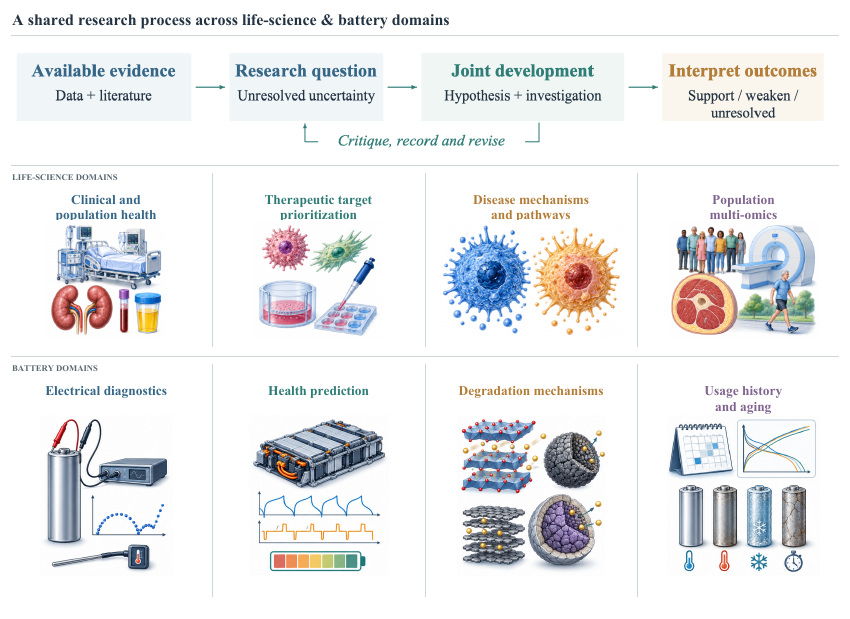}
\caption{\textbf{Shared research process across life-science and battery domains.} The process strip summarizes how evidence becomes a question, investigation, and evidence-bounded interpretation with iterative critique. The two rows show the four life-science domains and four battery domains used in the experiments.}
\label{fig:domain-research-process}
\end{figure}

\paragraph{Research domains.}
Our main evaluation of DiscoveryRSI covers four life-science domains: clinical and population health, therapeutic target prioritization, disease mechanisms, and population multi-omics. Inputs include the HCC clinical cohort, MIMIC-IV \citep{mimiciv2023}, eICU \citep{eicu2018}, UK Biobank \citep{ukbiobank2015}, and molecular studies from GEO and ArrayExpress \citep{geo2002,arrayexpress2021}. 
To test generalization beyond life science, we add a battery setting with four tasks: electrical diagnostics, health prediction, degradation mechanisms, and usage history and aging, specified in \Cref{fig:domain-research-process}. Agents inspect data and literature, formulate hypotheses, and develop investigation plans specifying comparisons, measurements, and interpretation criteria.

\paragraph{Hypothesis-credit training.}
We perform full-parameter fine-tuning of DiscoveryRSI-27B with a 256K-token context window using hypothesis-credit-weighted supervision, using Qwen3.8-27B as initialization. From 15,792 collected research trajectories, attribution, filtering, and tokenization yield 2,397 training trajectories: 634 Lead, 861 Generation, and 902 Evolution histories. The training sequences contain 291.12 million tokens, of which 6.18 million (2.12\%) receive direct supervision. We train for one epoch using Adam with a learning rate of $5\times10^{-6}$, a global batch size of 16, and BF16 precision. Detailed trajectory accounting, credit allocation, and token masking appear in \Cref{app:training}.

\paragraph{Evaluation.}
We examine four questions: whether hypothesis quality improves across research rounds, how research behavior changes, whether human reviewers recognize useful revisions, and whether hypotheses inspire subsequent research. Automated quality evaluation covers 241 research questions across the four domains using GPT-5.6 Sol as the judge. Behavioral analysis examines six campaigns of 20 episodes per evaluated model variant. Human assessment covers 292 proposals reviewed by 27 participants with biomedical or clinical research backgrounds under three senior supervisors. Proposals are presented in randomized order with model identities and automated scores hidden (\Cref{sec:expert-assessment}). Literature-based evaluation uses a shared reference corpus with model-specific temporal cutoffs. 

\section{Results}
\label{sec:results}

\subsection{RQ1: Does learning from research histories improve hypothesis quality?}

\begin{figure}[!htbp]
    \centering
    \includegraphics[width=1\linewidth]{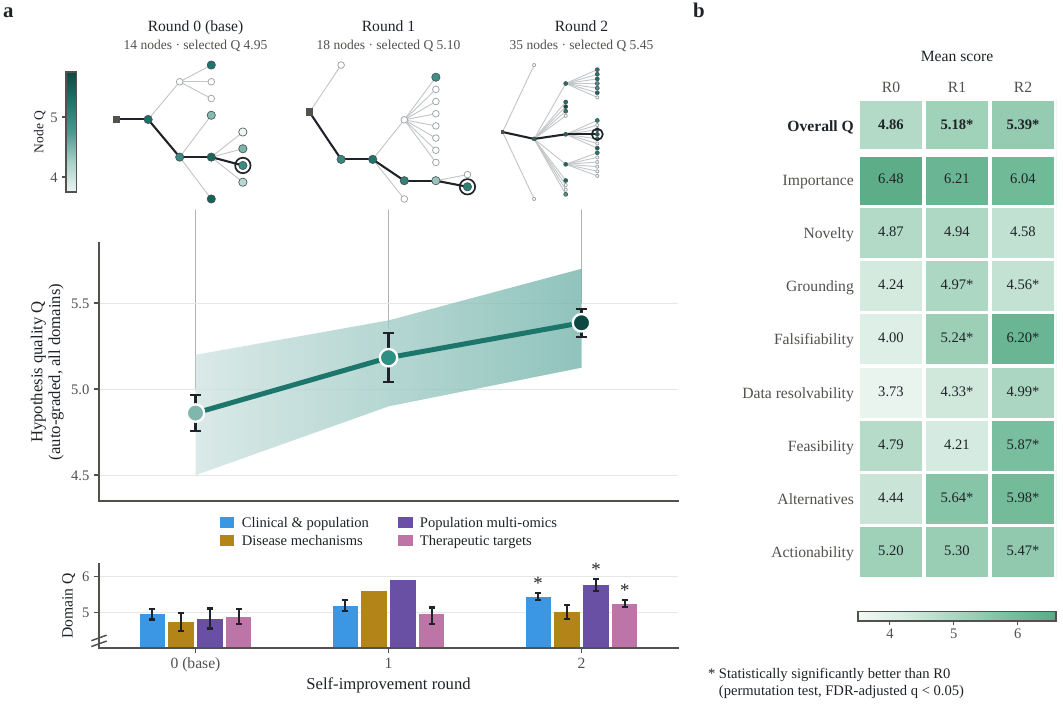}
    \caption{\textbf{Automated hypothesis quality across self-improvement rounds.} \textbf{(a)} Top: example research tree from each round; color shows weighted quality across eight evaluation axes. Middle: mean hypothesis quality per round. Error bars show 95\% confidence intervals and shading shows the interquartile range. Bottom: mean quality per domain with 95\% confidence intervals. \textbf{(b)} Mean and axis-specific scores. All scores are produced by the automated grader using the weights in \Cref{app:dfm-protocol}.}
    \label{fig:analysis1}
\end{figure}

\paragraph{Hypothesis quality rises with each round of self-improvement.} Mean automated hypothesis quality increases from 4.86 for the base model to 5.18 after the first research round and 5.39 after the second (\Cref{fig:analysis1}). The second-round model exceeds the base model in all four domains, with gains ranging from 0.28 points in disease mechanisms to 0.94 points in population multi-omics. The additional gain decreases from 0.32 in the first round to 0.21 in the second.

\paragraph{Improvements concentrate on the specification of testable investigations.} Falsifiability rises from 4.00 to 6.20, consideration of alternative explanations from 4.44 to 5.98, data resolvability from 3.73 to 4.99, and feasibility from 4.79 to 5.87. These changes indicate more explicit tests, better-defined evidence requirements, and clearer interpretation of possible outcomes. Gains are uneven: importance declines from 6.48 to 6.04 and novelty from 4.87 to 4.58. The observed improvement therefore concerns investigation design more strongly than the perceived significance or originality of research questions.

\subsection{RQ2: How does DiscoveryRSI change research behavior?}

\begin{figure}[!htbp]
    \centering
    \includegraphics[width=1\linewidth]{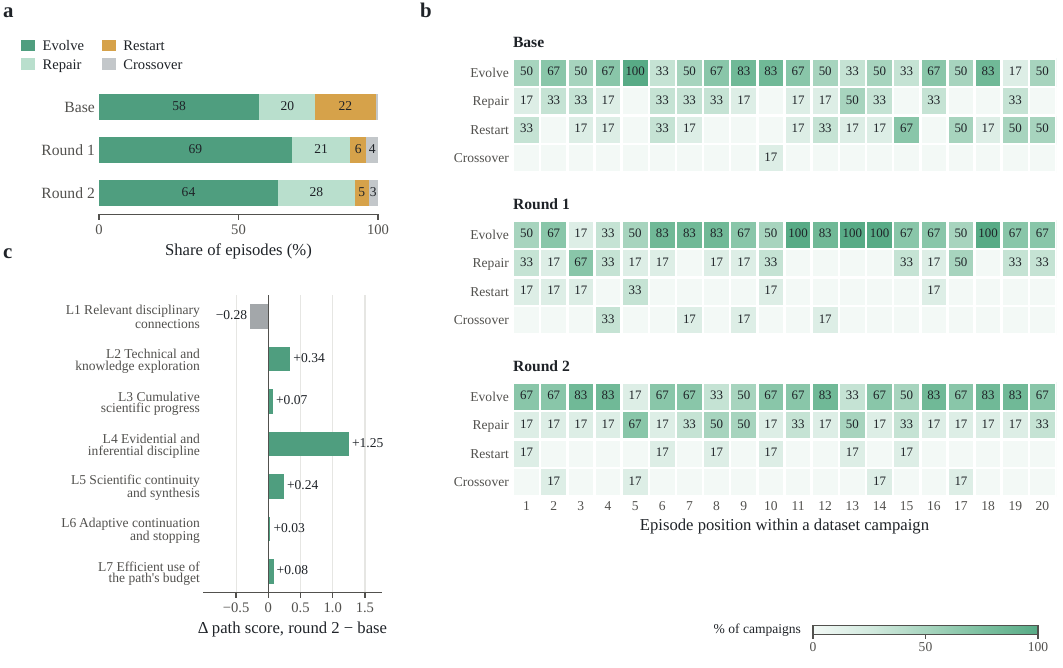}
    \caption{\textbf{Changes in research behavior across rounds.} \textbf{(a)} Share of each action across all episodes in a round. \textbf{(b)} Share of campaigns taking each action at each episode position, sampled from six campaigns per round; blank cells indicate 0\%. \textbf{(c)} Change in development-path scores from the base model to round 2.}
    \label{fig:analysis2}
\end{figure}

\paragraph{DiscoveryRSI allocates more effort to developing existing hypotheses.} Evolve and repair operations increase from 77.5\% of episodes for the base model to 90.0\% after the first round and 91.7\% after the second. Restarts decrease from 21.7\% to 5.8\% and 5.0\%, respectively (\Cref{fig:analysis2}). The shift is especially pronounced in episodes 11--20, where restarts fall from approximately 32\% for the base model to 3\% after the second round.

\textbf{Hypothesis development paths become more evidence-driven.} Development-path scores increase most for evidential and inferential discipline ($+1.25$), followed by technical and knowledge exploration ($+0.34$) and scientific continuity and synthesis ($+0.24$). Changes in cumulative progress, stopping decisions, and budget efficiency remain below 0.1 points, while disciplinary connections decline by 0.28. Together, these results suggest more sustained refinement and stronger evidential discipline, with limited improvement in broader exploration or resource allocation. Operation frequencies describe the research strategy; their value depends on the quality of the resulting revisions.

\subsection{RQ3: Do human reviewers recognize improvements in the hypotheses?}
\label{sec:human-evaluation-results}

\begin{figure}[!htbp]
\centering
\includegraphics[width=\linewidth]{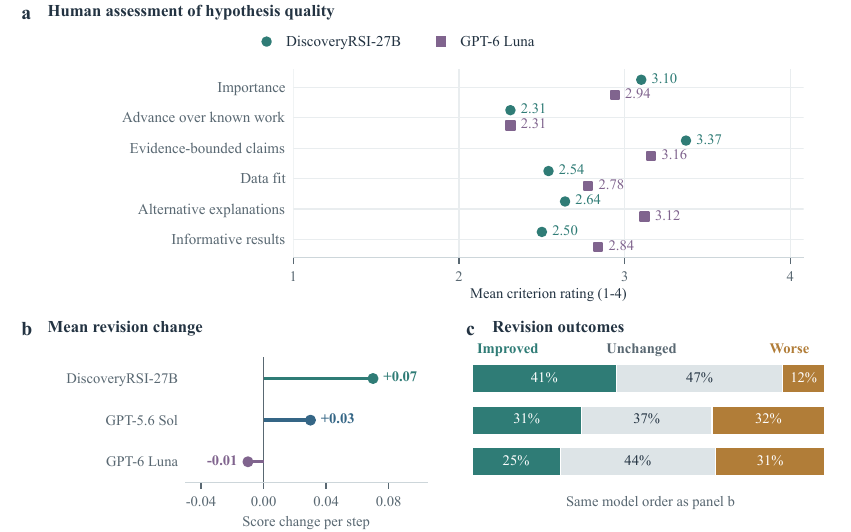}
\caption{\textbf{Human assessment of hypothesis quality and revision progress.} \textbf{(a)} Mean ratings across six criteria on a 1--4 scale. \textbf{(b)} Mean score change per revision. \textbf{(c)} Fractions of improved, unchanged, and worse revisions, in the same model order as panel (b).}
\label{fig:human-quality-progress}
\end{figure}

Human assessment identifies a favorable balance of revision outcomes for DiscoveryRSI-27B. Within the assessed research trees, 41\% of revisions improve the mean human score, 47\% leave it unchanged, and 12\% reduce it. The corresponding proportions are 31\%, 37\%, and 32\% for GPT-5.6 Sol, and 25\%, 44\%, and 31\% for GPT-6 Luna. Mean score change per revision is $+0.07$ for DiscoveryRSI-27B, compared with $+0.03$ and $-0.01$ for the two comparison models (\Cref{fig:human-quality-progress}). These results measure progress within research trees and provide complementary evidence to the comparisons across research rounds.

Absolute quality ratings show a mixed profile. DiscoveryRSI-27B scores higher than GPT-6 Luna on importance (3.10 versus 2.94) and plausible, evidence-bounded claims (3.37 versus 3.16). Both receive 2.31 for advance over known work, while GPT-6 Luna scores higher on data fit, alternative explanations, and informative results. Thus, the stronger revision outcomes coexist with remaining weaknesses in individual proposals.

\begin{figure}[!htbp]
\centering
\includegraphics[width=\linewidth]{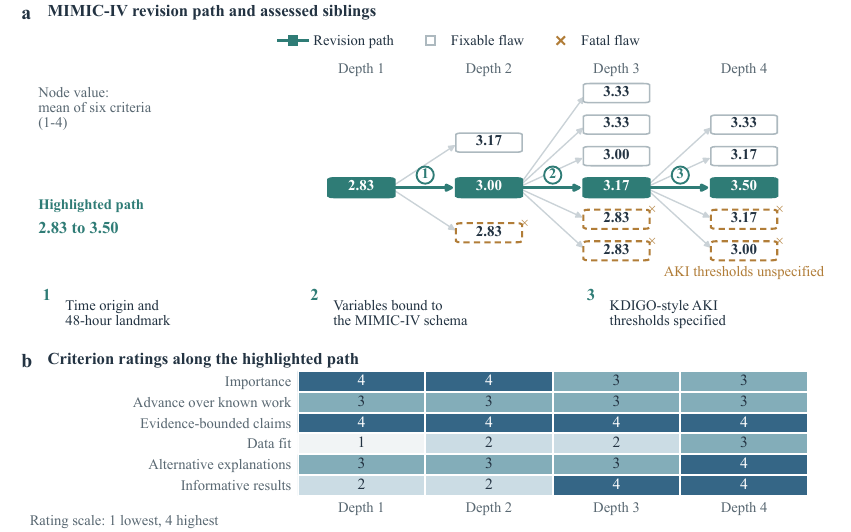}
\caption{\textbf{Design changes and human ratings along a revision path.} \textbf{(a)} A MIMIC-IV path and assessed siblings; node values are six-criterion means on a 1--4 scale. Numbered changes are explained below the tree. Dashed ochre nodes mark fatal flaws; the two final fatal siblings lack specified AKI thresholds. \textbf{(b)} Criterion ratings along the highlighted path.}
\label{fig:human-revision-case}
\end{figure}

\paragraph{A revision path in MIMIC-IV.}
A MIMIC-IV development path illustrates how useful revisions improve an investigation (\Cref{fig:human-revision-case}). Successive versions clarify the time origin and 48-hour landmark, bind variables to available data fields, and specify acute kidney injury thresholds. The mean of six human-rated criteria increases from 2.83 to 3.50; data fit and informative results each improve by two points. This example connects revision gains to concrete design changes. Other branches retain consequential flaws, including two depth-four siblings with unspecified acute kidney injury thresholds, underscoring the need to evaluate each resulting proposal.

\subsection{RQ4: Do the hypotheses inspire subsequent research?}
\label{sec:reference-actionability}

We define ``inspiration'' as a concrete connection between a generated hypothesis and subsequent literature. A publication after the recorded cutoff must provide an experimental design, comparator, measurement, or testable evidence gap relevant to developing or testing the hypothesis. 

\begin{figure}[!htbp]
\centering
\includegraphics[width=0.94\linewidth]{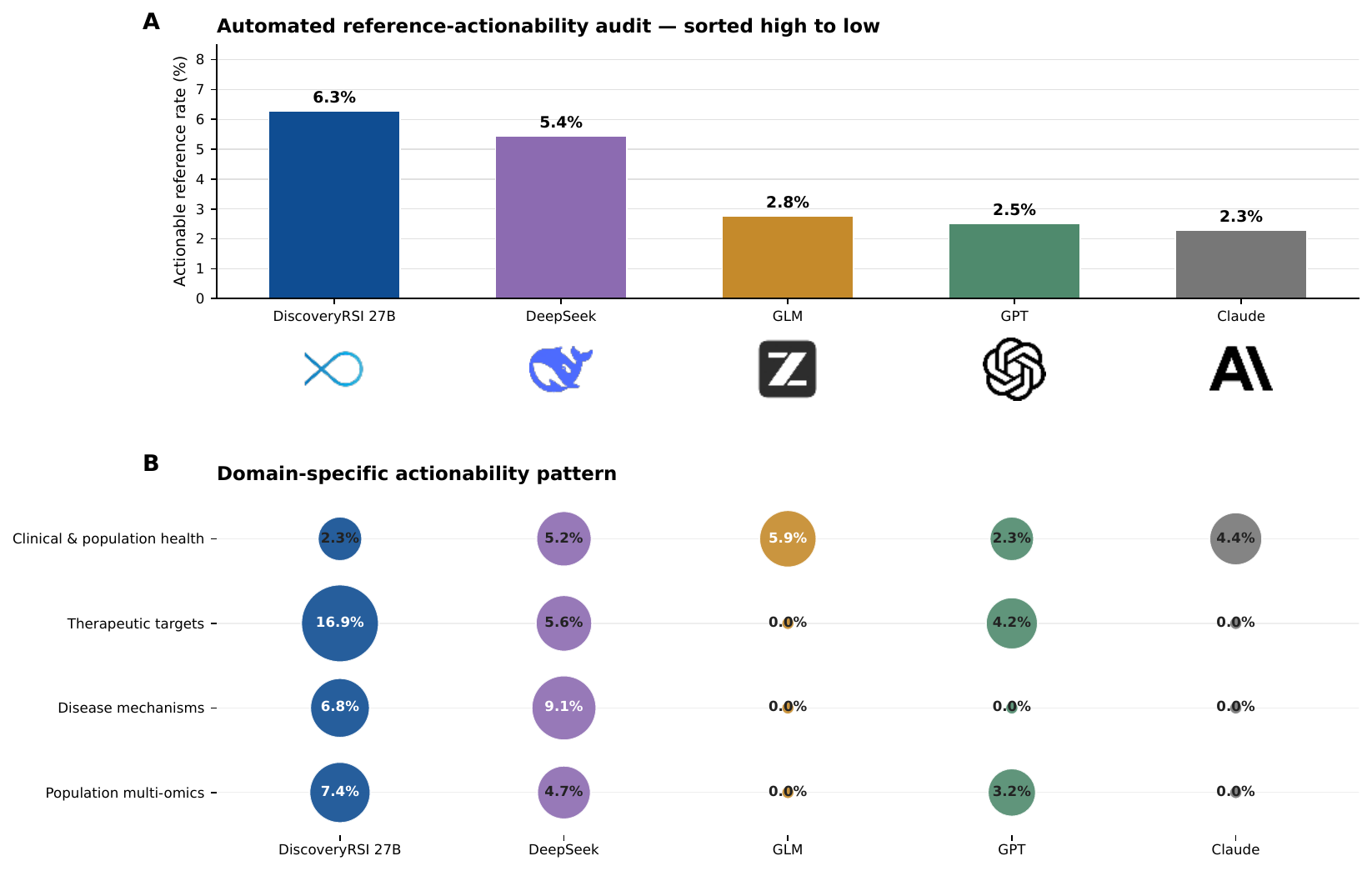}
\caption{\textbf{Hypothesis inspiration measured through subsequent literature.} The upper panel orders model groups by the proportion of generated hypothesis versions meeting the inspiration criterion. The lower panel reports the same rate across four domains. DiscoveryRSI-27B pools two production runs; Qwen original is omitted. The figure's reference-actionability labels denote this retrospective inspiration measure.}
\label{fig:expert-actionability}
\end{figure}

Across the two pooled DiscoveryRSI production runs, 74 of 1,176 hypothesis versions meet this criterion, giving an inspiration rate of 6.3\%. The corresponding rates are 5.4\% for DeepSeek (14/257), 2.8\% for GLM (1/36), 2.5\% for GPT (59/2,341), and 2.3\% for Claude (3/131). DiscoveryRSI reaches 16.9\% in therapeutic targets, 7.4\% in population multi-omics, 6.8\% in disease mechanisms, and 2.3\% in clinical and population health (\Cref{fig:expert-actionability}).

These descriptive results indicate that DiscoveryRSI hypotheses more frequently connect to concrete research directions in the retrieved literature. Group sizes, domain composition, temporal cutoffs, and retrieval coverage differ, and multiple versions may belong to the same research family. Prospective tracking is needed to determine whether researchers adopt these hypotheses and pursue the proposed investigations.

\paragraph{Literature comparison protocol.}
We construct a shared reference corpus through 420 backward and forward citation traversals, yielding 17,790 citation edges and 16,440 resolved publications. References are filtered using each model group's recorded cutoff. The inspiration rate is the proportion of generated hypothesis versions receiving a positive automated judgment under the above definition:
\begin{equation}
R_{\mathrm{insp}}=\frac{\sum_h a_h}{|\mathcal H|},
\label{eq:reference-actionability-rate}
\end{equation}
where $a_h$ indicates a positive judgment and $\mathcal H$ contains all generated hypothesis versions in the group. Corpus construction, cutoff handling, retrieval, and auditing are detailed in \Cref{app:reference-actionability}.

\paragraph{Why does Claude Opus 5 have a lower inspiration rate?}
We attribute part of Claude Opus 5's lower measured rate to retrieval coverage and proposal format. Only 15 of its 131 proposals pass the retrieval gate (11.5\%), compared with 355 of 1,176 for DiscoveryRSI-27B (30.2\%). Among these retrieved proposals, however, the positive-judgment fractions are similar: 20.0\% and 20.8\%, respectively. Claude's median proposal length is 41,134 characters, while retrieval uses the first 6,000 characters and the semantic audit sees the first 2,500. Provenance and revision notes can therefore displace scientific content from the evaluated excerpt. In a diagnostic using proposal titles as queries while retaining the corpus, scoring method, date rule, and overlap threshold, Claude's retrieval passage increases from 15 to 95 proposals. This sensitivity supports a format-related retrieval explanation, but does not establish additional matches meeting the inspiration criterion. 

\subsection{RQ5: How well does DiscoveryRSI transfer beyond life sciences?}

To test transfer beyond life science, we perform an additional battery domain evaluation covering electrical diagnostics, health prediction, degradation mechanisms, and usage history and aging. The evidence pack covers capacity and impedance aging measurements \citep{luh2024battery}, impedance-based health prediction \citep{zhang2020impedance,jones2022impedance}, degradation-mode analysis \citep{li2025degradation,kirkaldy2024ageing}, and path-dependent aging experiments \citep{raj2020path}. Investigations separate measured quantities from fitted degradation labels and specify cell, condition, and protocol splits. The same Lead--Generation--Evolution--Critique--Compiler loop is used, with battery-specific checks for operating conditions, sensing, physical plausibility, and competing explanations.

The battery domain applies the five-level DFM-Bench evaluation to 200 episodes and 351 generated hypotheses, with three AI reviews per hypothesis. The current panel reports a mean hypothesis quality of 5.61 on 165 complete scores, substantive progress on 214 of 366 transitions (58.47\%), a mean complete-trajectory score of 81.52 across 152 paths, and a research round score of 71.25. 20 contribution families were identified, with 7 carrying diagnostic or research evidence. 
\section{Related Work}
\label{sec:related-work}

\paragraph{Discovery agents and evaluation.}
Discovery agents increasingly support hypothesis generation, experimental planning, and computational investigation. Co-Scientist develops hypotheses through multi-agent refinement \citep{coscientist}, AI Scientist-v2 connects research ideation with agentic experimental search \citep{aiscientistv2}, and Biomni integrates biomedical tools and data \citep{biomni}. Discovery Foundation Models frame problem formation and evidence-grounded revision as learnable capabilities \citep{dfm}, while ScienceIDE constructs environments for learning from scientific workflows \citep{scienceide}. Evaluation spans scientific ideation in AgentIdeaBench \citep{agentideabench}, data-driven discovery in DiscoveryBench and BLADE \citep{discoverybench,blade}, executable investigations in ScienceAgentBench \citep{scienceagentbench}, and research-process assessment in OEB \citep{oeb}. DiscoveryRSI connects hypothesis development to subsequent model learning, while DFM-Bench makes hypotheses, revisions, development paths, research runs, and contribution families explicit evaluation objects.

\paragraph{Learning from evaluated research experience.}
Reward-guided training provides a foundation for converting generated experience into model updates. RAFT selects highly rated samples for fine-tuning \citep{raft}, ReST-EM alternates generation and feedback-based filtering \citep{restem}, and advantage-weighted regression learns through weighted likelihood objectives \citep{awr}. SimpleTES further studies learning from scientific exploration trajectories \citep{simpletes}. These approaches motivate the use of assessed research histories as supervision. DiscoveryRSI addresses the attribution problem within multi-agent hypothesis development: an assessment concerns a scientific proposal, whereas learning targets belong to the histories that contributed to it. Hypothesis credit optimization links these levels through documented contributions, bounded credit allocation, and supervision restricted to supported assistant actions.

\paragraph{Persistent and recursive agent improvement.}
Reflexion, GEPA, and ACE retain experience through verbal memory, prompt evolution, and contextual playbooks \citep{reflexion,gepa,ace}; harness optimization extends adaptation to agent workflows and implementations \citep{selfharness,agenticharness}. Darwin G\"odel Machine and Hyperagents study persistent program-level improvement \citep{dgm,hyperagents}, while MASS couples multi-agent workflow optimization with learning from selected executions \citep{mass}. ScienceBuddy connects harness refinement with model learning \citep{sciencebuddy}, and RSI-Anything uses predictive world models to guide interaction and retained system updates \citep{rsianything}. DiscoveryRSI 
focuses on how experience accumulated while developing scientific questions can improve later investigations.

\Needspace{5\baselineskip}
\section{Conclusion}
\label{sec:conclusion}

DiscoveryRSI connects hypothesis development with learning from documented research contributions. The present work reaches concrete hypotheses and compiled investigation plans. The next step is to determine which proposals merit investigation, using solvers and frameworks such as RSI-Anything \citep{rsianything} to execute, assess, and refine them. Closing the discovery loop further requires wet-lab validation and methods that turn supportive, negative, and inconclusive outcomes into improved research decisions. Extending hypothesis discovery and DFM-Bench to additional scientific domains will help establish how these capabilities generalize. Together, these directions advance the DFM vision \citep{dfm} of agents that formulate worthwhile questions, investigate them, and learn from the resulting evidence.

\clearpage
\bibliography{discovery-refs}
\clearpage
\linespread{1.10}\selectfont
\appendix
\crefname{appendix}{Appendix}{Appendices}
\Crefname{appendix}{Appendix}{Appendices}
\crefalias{section}{appendix}
\crefalias{subsection}{appendix}
\titleformat{\section}{\sffamily\bfseries\fontsize{13}{15}\selectfont}{Appendix \thesection.}{0.5em}{}
\pdfbookmark[0]{DiscoveryRSI: Supplementary Methods and Evaluation}{supplement-title}
\begin{center}
{\sffamily\bfseries\Large DiscoveryRSI\par}
\vspace{0.4em}
{\sffamily\bfseries\large Supplementary Methods and Evaluation\par}
\end{center}
\noindent This supplement details the research workflow, hypothesis credit optimization, training implementation, and evaluation protocols supporting the main report. Appendices A--C describe the system and learning procedure; Appendices D--F describe automated assessment, human review, and literature-inspiration evaluation. Notation is defined where it is used. References to main-text figures and equations retain their original numbering.

\subsection*{Appendix Contents}
\newcommand{\suppentry}[2]{\noindent\hyperref[#1]{\textbf{Appendix \ref*{#1}. #2}}\nobreak\dotfill\pageref*{#1}\par}
\newcommand{\suppsubentry}[2]{\noindent\hspace*{1em}\hyperref[#1]{\ref*{#1}. #2}\nobreak\dotfill\pageref*{#1}\par}
\begingroup\small\setlength{\parskip}{1pt}
\suppentry{app:linked-records}{Research Workflow and Linked Records}
\suppsubentry{supp:a-1}{Research Objects and Agent Roles}
\suppsubentry{supp:a-2}{Record Types and Provenance}
\suppsubentry{supp:a-3}{From Contributions to Executable Investigations}
\suppsubentry{supp:a-4}{Harness Replay and Rediscovery}
\suppentry{app:credit}{Hypothesis Credit Optimization}
\suppsubentry{supp:b-1}{Pairwise Preference Assessment}
\suppsubentry{supp:b-2}{Hypothesis Weights}
\suppsubentry{supp:b-3}{Contributor Shares and Provenance}
\suppsubentry{supp:b-4}{Scientific-Family Credit Limits}
\suppsubentry{supp:b-5}{Objective Balancing}
\suppsubentry{supp:b-6}{Allocation Parameters}
\suppsubentry{supp:b-7}{Credit Conservation}
\suppentry{app:training}{Training Data and Implementation}
\suppsubentry{supp:c-1}{Trajectory Collection and Selection}
\suppsubentry{supp:c-2}{Supervision and Optimization}
\suppsubentry{supp:c-3}{Training Configuration}
\suppsubentry{supp:c-4}{Serialization and Target Selection}
\suppsubentry{supp:c-5}{Windowing: Preserved Targets and Conditional Equivalence}

\suppentry{app:dfm-protocol}{DFM-Bench Evaluation Protocol}
\suppsubentry{supp:d-1}{Hypothesis Quality and Automated Review}
\suppsubentry{supp:d-2}{Revisions, Development Paths, and Research Runs}
\suppsubentry{supp:d-3}{Scientific Contribution, Delivery, and Calibration}
\suppsubentry{supp:d-4}{Development-Path and Research-Run Rubrics}
\suppsubentry{supp:d-5}{Assessment Records and Missing Evidence}
\suppsubentry{supp:d-6}{Benchmark Coverage Definitions}
\suppentry{app:human-assessment}{Human Assessment Protocol}
\suppsubentry{supp:e-1}{Review Cohort and Presentation}
\suppsubentry{supp:e-2}{Rating Scales and Recommendations}
\suppsubentry{supp:e-3}{Revision Outcomes and Interpretation}
\suppentry{app:reference-actionability}{Literature-Inspiration Evaluation}
\suppsubentry{supp:f-1}{Reference Corpus and Temporal Boundaries}
\suppsubentry{supp:f-2}{Retrieval and Automated Judgment}
\suppsubentry{supp:f-3}{Population Accounting}
\suppsubentry{supp:f-4}{Claude Opus 5: Retrieval Sensitivity}
\endgroup
\clearpage
\section{Research Workflow and Linked Records}
\label{app:linked-records}
DiscoveryRSI develops hypotheses comprising a scientific claim, investigation plan, decision criteria, and alternative explanations. This appendix describes the roles and records connecting hypothesis development to task delivery and subsequent learning.

\subsection{Research Objects and Agent Roles}
\label{supp:a-1}
\label{app:workflow-objects}
A \emph{hypothesis version} is one recorded proposal; a \emph{revision} links it to its predecessor. Crossover can introduce several parents, so the \emph{research tree} is generally a directed acyclic graph. A \emph{development path} follows ordered hypothesis versions; a \emph{research run} contains the complete graph and its decisions. An \emph{agent trajectory} records one agent invocation, including instructions, actions, and observations. Development paths and agent trajectories are distinct evaluation and training units.

Lead chooses directions, parents, operations, and selections. Generation proposes hypotheses; Evolution develops or repairs them. Critique provides blinded preferences, Progress assesses revisions against ancestry, and Compiler translates selected plans into Harbor tasks. Evolve and repair retain an existing direction, crossover combines directions, and restart opens a new question. Shared context can inform a restart without creating parentage.

\subsection{Record Types and Provenance}
\label{supp:a-2}
\label{app:record-types}
The record layer links hypotheses to evidence, data, contributions, and execution. Ancestry records scientific development; evidence reuse and execution dependencies are separate relations. Table~\ref{tab:linked-records} summarizes the retained records.

\begin{table}[!htbp]
\small\centering
\caption{\textbf{Persistent records supporting research and recursive improvement.} Files retain material; explicit references retain provenance and relationships.}
\label{tab:linked-records}
\begin{tabularx}{\linewidth}{@{}p{3.0cm}YY@{}}
\toprule
Record type & Contents & Recorded relationships \\
\midrule
Hypothesis versions & Claim, plan, criteria, alternatives, and parents. & Revision ancestry; restart context recorded separately. \\[1mm]
Literature and evidence & Sources, excerpts, and retrieval metadata. & Origin, dates, versions, and content hashes. \\[1mm]
Data dependencies & Files and inputs required for investigation. & Bindings to hypotheses, jobs, and tasks. \\[1mm]
Supporting artifacts & Scripts, results, plots, and evidence notes. & Explicit hypothesis attachments with paths and hashes. \\[1mm]
Drafts and notes & Working proposals and continuation notes. & Agent ownership and recorded publication links. \\[1mm]
Scientific assessments & Validity, preferences, and Progress judgments. & Assessed objects, comparison pairs, and ancestors. \\[1mm]
Contributions and decisions & Assignments, roles, and publication events. & Agent, episode, attempt, and work identifiers. \\[1mm]
Execution and delivery & Jobs, outputs, packages, and validation results. & Input dependencies and produced artifacts. \\
\bottomrule
\end{tabularx}
\end{table}

Registry entries and filesystem artifacts implement these relationships. Source records retain retrieval time and available publication metadata separately. A provenance link establishes recorded use or production; scientific assessment determines how the material supports or limits a claim.

\subsection{From Contributions to Executable Investigations}
\label{supp:a-3}
\label{app:record-attribution}
Credit preparation links an assessed hypothesis to producing, supporting, and coordinating agent trajectories. Publication identity, original drafts, artifact hashes, and event order distinguish original work from later quotation or reuse. These links supply the contributor pools in \Cref{eq:contributor-share}. Unresolved associations receive no allocated share.

For example, Generation may propose a comparison, Evolution may revise it after inspecting evidence, and Lead may supply the assignment that enabled that work. Their records can identify supported actions across several trajectories. Contribution allocation determines each trajectory's weight; target selection determines which of its assistant outputs receive supervision (\Cref{app:credit,app:training}).

A selected computational investigation is compiled into a Harbor task containing its inputs, environment, required outputs, and verification checks. Compilation preserves the population or biological system, comparisons, discriminating outcomes, and decision criteria. Unavailable measurements and wet-lab experiments remain explicit dependencies. Compilation and readiness checks assess the computational package; completed execution and scientific assessment establish what evidence the investigation supplies.

\subsection{Harness Replay and Rediscovery}
\label{supp:a-4}
\label{app:harness-design}
\label{app:harness-substrate}
DiscoveryRSI uses Codex and Claude Code for sessions, context management, filesystem access, and tool execution \citep{codexharness,claudecode}. Research adapters connect these facilities to agent roles, shared records, computation, and task delivery. Harness changes concern coordination, evidence access, review timing, and research allocation.

\paragraph{Diagnosis and replay.}
\label{app:harness-diagnosis}
\label{app:recursive-implementation}
Diagnostic runs hold the checkpoint and harness fixed. Linked decisions and execution records distinguish scientific-plan failures from tool failures, lost artifacts, and premature closure. Candidate repairs address evidence reuse, handoffs, retries, or completion checks; these examples do not constitute ablation results.

Replay preserves observations, tool access, and memory at the restored point. Alternatives require recorded continuations or new execution from a compatible state; ancestry alone supplies no counterfactual outcomes. This use of experience relates to Dream-RSI's exploration-policy perspective \citep{dreamrsi}.

\paragraph{Acceptance and renewed investigation.}
Harness comparisons hold the checkpoint, inputs, data access, and resource limits fixed. Model and harness candidates are assessed separately and jointly before deployment on fresh investigations (Algorithm~\ref{alg:dsi-loop}).

\section{Hypothesis Credit Optimization}
\label{app:credit}
Hypothesis credit optimization converts scientific assessments into weighted supervision for documented research contributions. The procedure first estimates relative hypothesis preference, then assigns hypothesis weights, allocates contributor shares, limits credit within scientific families, and balances the training objective. Bayesian preference inference and contribution allocation are distinct steps.

\subsection{Pairwise Preference Assessment}
\label{supp:b-1}
\label{app:trueskill}

The preference model summarizes blinded pairwise Critique outcomes under a common scientific rubric. Its posterior describes relative preference strength; contributor attribution is determined separately from provenance.

The runtime collects blinded pairwise Critique outcomes: preference for A or B, a tie, or insufficient evidence. To derive the preference posterior, treat each versioned hypothesis as the rated entity in TrueSkill \citep{trueskill}. Its latent quantity $z_h$ represents preference strength under the scientific critique rubric, rather than an actor's personal skill. For one compatible comparison population, take Gaussian priors and noisy comparison performances:
\begin{equation}
 z_h\sim\mathcal N(\mu_{h,0},\sigma_{h,0}^2),\qquad
 t_{h,e}=z_h+\epsilon_{h,e},\qquad
 \epsilon_{h,e}\sim\mathcal N(0,\beta^2).
 \label{eq:trueskill-generative}
\end{equation}
For comparison $e=(a,b,o_e)$, the observed outcome depends on $D_e=t_{a,e}-t_{b,e}$: A is preferred when $D_e>\varepsilon$, B when $D_e<-\varepsilon$, and a tie when $|D_e|\leq\varepsilon$. Here $\beta$ models performance noise and $\varepsilon$ is the draw margin. With $\Delta z=z_a-z_b$ and standard normal CDF $\Phi$, integrating out performance noise gives
\begin{align}
 P(A\mid z_a,z_b)&=\Phi\!\left(\frac{\Delta z-\varepsilon}{\sqrt{2}\beta}\right),\label{eq:preference-likelihood}\\
 P(B\mid z_a,z_b)&=\Phi\!\left(\frac{-\Delta z-\varepsilon}{\sqrt{2}\beta}\right),\nonumber\\
 P(\mathrm{tie}\mid z_a,z_b)&=
 \Phi\!\left(\frac{\varepsilon-\Delta z}{\sqrt{2}\beta}\right)
 -\Phi\!\left(\frac{-\varepsilon-\Delta z}{\sqrt{2}\beta}\right).\nonumber
\end{align}
Insufficient evidence is an abstention and supplies no preference-likelihood factor. Under the comparison model's conditional-independence assumption, Bayes' rule yields the joint posterior
\begin{equation}
 p(\mathbf z\mid\mathcal D_{\rm cmp})=
 \frac{1}{Z}\prod_h\mathcal N(z_h;\mu_{h,0},\sigma_{h,0}^2)
 \prod_{e\in\mathcal D_{\rm cmp}}P(o_e\mid z_{a_e},z_{b_e}),
 \label{eq:joint-preference-posterior}
\end{equation}
where $Z$ is the evidence normalizer. Pairwise factors couple hypotheses, so this joint posterior is generally neither Gaussian nor factorized. TrueSkill uses approximate message passing and Gaussian moment matching to retain tractable marginal beliefs. Stored online updates summarize the evidence available at their update time; they are reused without claiming an exact batch re-inference of the complete comparison graph.

\paragraph{How a comparison changes the posterior.}
For a single decisive A-preferred comparison with independent Gaussian incoming beliefs, let
\begin{equation}
 m=\mu_a-\mu_b,\quad
 c^2=\sigma_a^2+\sigma_b^2+2\beta^2,\quad
 u=\frac{m-\varepsilon}{c},\quad
 V(u)=\frac{\phi(u)}{\Phi(u)},\quad W(u)=V(u)\bigl[V(u)+u\bigr],
 \label{eq:trueskill-moments}
\end{equation}
with standard normal density $\phi$. Before observing the outcome, $D\sim\mathcal N(m,c^2)$. Conditioning on $D>\varepsilon$ truncates this Gaussian, giving $\E[D\mid A]=m+cV(u)$ and $\operatorname{Var}(D\mid A)=c^2[1-W(u)]$. Gaussian conditioning transfers these moments to the latent hypotheses:
\begin{align}
 \mu_a'&=\mu_a+\frac{\sigma_a^2}{c}V(u),&
 \mu_b'&=\mu_b-\frac{\sigma_b^2}{c}V(u),\label{eq:trueskill-mean-update}\\
 (\sigma_a')^2&=\sigma_a^2\!\left[1-\frac{\sigma_a^2}{c^2}W(u)\right],&
 (\sigma_b')^2&=\sigma_b^2\!\left[1-\frac{\sigma_b^2}{c^2}W(u)\right].
 \label{eq:trueskill-variance-update}
\end{align}
These are the exact marginal first two moments for this single Gaussian-input comparison; retaining Gaussian beliefs is the approximation. A tie instead uses a two-sided truncated Gaussian. In a larger or sequential comparison graph, message schedules and Gaussian projection determine the stored approximation. Versioned runtime priors, noise, draw margin, and any dynamics variance remain part of the rating provenance; the static derivation does not silently replace those settings.

The stored marginals summarize relative preference strength within compatible source-snapshot, dataset, rubric, and connected comparison strata. Reviewer bias, scientific uncertainty, and dependence between repeated critiques are not fully represented by the Gaussian variance. Separate ancestor-aware Progress judgments supply transition evidence.

\subsection{Hypothesis Weights}
\label{supp:b-2}
For a rated hypothesis, define a conservative ranking statistic and its tie-aware within-stratum rank:
\begin{align}
 s_h&=\mu_h-\kappa\sigma_h,\label{eq:conservative-rating}\\
 r_h&=\frac{\sum_{k\in\mathcal S}\mathbf 1[s_k<s_h]
 +\tfrac12\sum_{k\in\mathcal S}\mathbf 1[s_k=s_h]}{|\mathcal S|}.
 \label{eq:local-rank}
\end{align}
Under the Gaussian marginal approximation, $s_h$ is the $\Phi(-\kappa)$ posterior quantile; for $\kappa=2$, approximately $2.3\%$ of the modeled preference mass lies below it. This is a conservative statistic of rubric-conditioned preference. A singleton has rank $1/2$, supplying no relative ordering. Let $n_h$ be the effective comparison count. Two reliability factors damp the contribution of sparse or uncertain ratings:
\begin{equation}
 \rho_h^{\rm cov}=\frac{n_h}{n_h+\nu},\qquad
 \rho_h^{\rm unc}=\frac{1}{1+\sigma_h/\sigma_{\rm ref}}.
 \label{eq:rating-reliability}
\end{equation}
Let $v_h$ indicate a recorded valid assessment and at least one effective comparison, with no recorded marginal-repetition judgment. Let $p_h$ indicate a recorded substantive-progress or necessary-repair judgment. The bounded outcome transformation is given in \Cref{eq:hyp-weight}. Validity here means a coherent, assessable research design under the stored rubric, not experimental confirmation. The gate controls eligibility for this weak-supervision route. A missing rating or assessment remains an unknown outcome; zero assigned loss is not a negative scientific label. A valid hypothesis motivated by a supported null result remains eligible.

\begin{equation}
 q_h=v_h\min\!\left\{q_{\max},\;
 q_{\min}+(q_{\max}-q_{\min})r_h\rho_h^{\rm cov}\rho_h^{\rm unc}
 +\lambda_p p_h\right\}.
 \label{eq:hyp-weight}
\end{equation}

Equation~\eqref{eq:hyp-weight} is our bounded transformation of a preference posterior, not a quantity inferred by TrueSkill. The ranking statistic and uncertainty factor both penalize uncertainty; comparison count also correlates with posterior concentration. Their individual effects require ablation. A Progress bonus supplements the outcome weight without changing the TrueSkill posterior.

\subsection{Contributor Shares and Provenance}
\label{supp:b-3}
We construct an auditable contribution graph whose edges identify an agent invocation's relationship to a hypothesis. Direct publication requires matching publisher, episode, and creation-time provenance. Documented production or support can additionally be recovered from original prepublication proposal text, an exact draft-revision promotion record, or emitted supporting-file bytes whose hash and original path match the frozen candidate record. Lead coordination requires a contributing committed work assignment or publication/promotion record. Merely sharing an episode, appearing in an ancestry graph, or completing a worker invocation creates no credit edge. These links establish recorded association and production provenance; they do not establish sole authorship or the artifact's scientific correctness.

For hypothesis $h$, let $L_h$ contain verified coordination trajectories and $P_h$ contain direct production or documented supporting trajectories. Each trajectory belongs to at most one pool for a given hypothesis. With coordination fraction $\gamma$, define
\begin{equation}
 \alpha_{ih}=\begin{cases}
 \gamma/|L_h|,&i\in L_h,\\
 (1-\gamma)/|P_h|,&i\in P_h,\\
 0,&\text{otherwise}.
 \end{cases}
 \qquad \sum_i\alpha_{ih}\leq1.
 \label{eq:contributor-share}
\end{equation}
Only nonempty pools are evaluated. If a contributor pool is missing, its share remains unallocated rather than being transferred to a convenient recipient. A Lead-only publication with unresolved authorship therefore receives the coordination share alone. Conversely, documented Lead production can enter $P_h$: the rule follows contribution function rather than a fixed role label.

For example, one coordinating Lead, one Generation publisher, and one Evolution trajectory with documented supporting work receive fractions $0.2$, $0.4$, and $0.4$ under the implemented $\gamma=0.2$ policy. This allocation reaches work that need not register its own hypothesis. However, the fractions are a fixed responsibility prior, not a Bayesian posterior over actors or a counterfactual estimate of their marginal contribution. The TrueSkill marginal informs $q_h$ through our explicit transformation; provenance supplies $\alpha_{ih}$. Critique votes cannot independently certify the correctness of those same critique outputs, and successful compilation cannot certify scientific fidelity. Such roles require their own evidence before receiving direct imitation targets.

\subsection{Scientific-Family Credit Limits}
\label{supp:b-4}
Duplicate registrations do not receive additional scientific credit. We suppress exact duplicate proposal-credit events within a source snapshot and dataset. Local ancestry families $F_{\rm loc}(h)$ also receive a bounded outcome budget:
\begin{equation}
 \widetilde q_h=q_h\min\!\left\{1,
 \frac{C_{\rm loc}}{\sum_{k\in F_{\rm loc}(h)}q_k}\right\},\qquad
 u_{ih}=\alpha_{ih}\widetilde q_h.
 \label{eq:local-credit-budget}
\end{equation}
The local sum covers rated comparison nodes, including imported anchors when present; only generated hypotheses with contributor evidence allocate training credit. Global scientific families $F_{\rm glob}$ join recorded ancestry and compatible candidate or exact-proposal identities across snapshots. Their allocated credit is capped before forming a trajectory weight:
\begin{equation}
 d_F=\min\!\left\{1,\frac{C_{\rm glob}}{\sum_{h\in F}\sum_i u_{ih}}\right\}.
 \label{eq:global-credit-budget}
\end{equation}
The final allocation from hypothesis $h$ to trajectory $i$ is $b_{ih}=d_{F_{\rm glob}(h)}u_{ih}$, and the trajectory weight is $w_i=\min(W_{\max},\sum_h b_{ih})$. For a local family with zero total weight, the local cap factor is defined as one. For a zero-mass family, the cap factor is defined as one and allocated credit remains zero. These budgets limit the credit assigned to repeated versions of the same scientific question. They are pre-normalization credit constraints; subsequent domain and group balancing determines their relative influence in the final objective. Unresolved semantic duplication remains an audit limitation rather than an assumed solved problem.

\subsection{Objective Balancing}
\label{supp:b-5}
Let $g_i$ identify a conservative shared-memory group, closed over campaign memory, ancestry, resumed trajectories, and recovered duplicates; these balancing groups are distinct from the scientific credit families above. For domain $d$, let $\mathcal G_d$ contain groups with retained targets and let $\eta_d$ be its desired objective mass. The fixed trajectory coefficient and intended update are
\begin{equation}
 \omega_i=\frac{\eta_{d_i}}{|\mathcal G_{d_i}|}
 \frac{w_i}{\sum_{k:\,d_k=d_i,\,g_k=g_i}w_k},\qquad \sum_d\eta_d=1.
 \label{eq:balanced-credit}
\end{equation}
All sums in \Cref{eq:balanced-credit} concern retained trajectories with positive target mass. Coefficients and denominators are fixed from the frozen training release and are not renormalized within individual minibatches. This prevents transcript length alone from determining a trajectory's loss weight. The coefficients allocate objective mass, not guaranteed shares of gradient norm. Initialization and optimization settings are reported in \Cref{app:training}.

\subsection{Allocation Parameters}
\label{supp:b-6}
\begingroup
\small
\begin{longtable}{@{}p{0.39\linewidth}p{0.13\linewidth}p{0.42\linewidth}@{}}
\toprule
Quantity & Value & Function\\
\midrule
\endfirsthead
\toprule Quantity & Value & Function\\
\midrule
\endhead
\bottomrule
\endfoot
Conservative-rating factor $\kappa$ & $2$ & Penalize uncertain preference means\\
Coverage constant $\nu$ & $2$ & Damp sparsely compared hypotheses\\
Uncertainty reference $\sigma_{\rm ref}$ & $8.3333$ & Fixed rating-scale reference\\
Weight interval $[q_{\min},q_{\max}]$ & $[0.5,2]$ & Bound eligible outcome weights\\
Progress supplement $\lambda_p$ & $0.25$ & Use recorded progress evidence\\
Coordination fraction $\gamma$ & $0.2$ & Reserve coordination share\\
Local/global family budgets & $2,2$ & Limit repeated scientific credit\\
Trajectory cap $W_{\max}$ & $2$ & Bound aggregated trajectory weight\\
Positive local credit $c_{it}$ & $0.5$ & Uniform positive token eligibility\\
\bottomrule
\end{longtable}
\endgroup
These constants specify the implemented allocation recipe; they are not estimates of causal contribution or empirically optimal hyperparameters. The uniform local credit cancels in the per-trajectory mean. Domain balancing is separate from scientific-quality ranking; a percentile within one stratum is not a universal cross-domain utility scale.

\subsection{Credit Conservation}
\label{supp:b-7}
\begin{samepage}
\begin{proposition}[Pre-normalization credit conservation]
\label{prop:credit-conservation}
Fix deduplicated hypothesis-credit events and scientific-family assignments. Assume nonnegative hypothesis weights, positive budgets, and contributor shares satisfying $\sum_i\alpha_{ih}\leq1$. Under the local and global scaling rules,
\[
 \sum_i b_{ih}\leq\widetilde q_h,\qquad
 \sum_{h\in F}\sum_i b_{ih}\leq C_{\rm glob},\qquad
 0\leq w_i\leq W_{\max}
\]
for each global scientific family $F$. Missing contributor shares remain unallocated.
\end{proposition}
\end{samepage}

\paragraph{Proof.}
\label{app:credit-proof}
Fix deduplicated credit events and a partition into global scientific families, with nonnegative weights, positive budgets, and $0\leq\gamma\leq1$. Every cap factor belongs to $[0,1]$. The mutually exclusive contributor pools give $\sum_i\alpha_{ih}\leq1$, hence
\begin{equation}
 \sum_i b_{ih}=d_{F_{\rm glob}(h)}\widetilde q_h\sum_i\alpha_{ih}
 \leq\widetilde q_h.
\end{equation}
For a global family $F$, let $U_F=\sum_{h\in F,i}u_{ih}$. If $U_F>0$, then
\begin{equation}
 \sum_{h\in F,i}b_{ih}=\min(1,C_{\rm glob}/U_F)U_F
 =\min(U_F,C_{\rm glob})\leq C_{\rm glob}.
\end{equation}
If $U_F=0$, all its allocations vanish. Similarly, the local scaling gives $\sum_{h\in F_{\rm loc}}\widetilde q_h\leq C_{\rm loc}$ for each fixed local family. Nonnegativity and the trajectory cap yield $0\leq w_i\leq W_{\max}$. Missing contributor pools do not enter $\alpha_{ih}$ and their reserved shares remain unallocated. This proves \Cref{prop:credit-conservation}. The result holds before objective balancing; group normalization can change relative family influence and supplies no corresponding bound on gradient magnitude.

\section{Training Data and Implementation}
\label{app:training}
This appendix documents the data and optimization used for full-parameter fine-tuning of DiscoveryRSI-27B. An agent trajectory is one recorded agent invocation, rather than a path through the hypothesis tree. Only supported assistant outputs receive direct loss; other retained tokens provide context.

\subsection{Trajectory Collection and Selection}
\label{supp:c-1}

The source collection contains 15,792 research trajectories. Structural review identifies 13,270 reviewable agent trajectories, of which 13,269 pass export-integrity verification. Attribution identifies 2,432 agent trajectories with eligible supervision targets. Native-template tokenization retains 2,397 agent trajectories; 35 agent trajectories with unsupported mid-conversation developer messages remain excluded from the training export.

\begin{table}[!htbp]
\centering\small
\caption{\textbf{Trajectory and token accounting for hypothesis-credit training.} Training-sequence tokens include supervised targets and zero-loss context. Collected trajectories and retained training trajectories are counted separately.}
\label{tab:credit-training-data}
\setlength{\tabcolsep}{4pt}
\begin{tabular}{@{}lrrrr@{}}
\toprule
Role & Collected & Training & Sequence tokens & Supervised tokens \\
\midrule
Lead & 3,423 & 634 & 98,357,399 & 977,931 \\
Generation & 2,502 & 861 & 96,414,270 & 2,437,985 \\
Evolution & 2,310 & 902 & 96,349,331 & 2,761,689 \\
Critique & 4,988 & 0 & --- & 0 \\
Compiler & 2,056 & 0 & --- & 0 \\
Progress & 513 & 0 & --- & 0 \\
\midrule
Total & 15,792 & 2,397 & 291,121,000 & 6,177,605 \\
\bottomrule
\end{tabular}
\end{table}

Lead, Generation, and Evolution account for 26.4\%, 35.9\%, and 37.6\% of retained training trajectories, respectively. Critique, Compiler, and Progress records remain available as sources or context but receive no direct supervision. Exclusion from supervision indicates insufficient support under the attribution policy and does not imply that a role or trajectory lacks scientific value.

\subsection{Supervision and Optimization}
\label{supp:c-2}

Scientific assessments and documented contributions determine fixed trajectory weights. Supervision targets supported assistant responses and tool-call arguments. Tool observations, failed actions, repeated context, and unsupported actions receive zero direct loss. For trajectory $i$, let $\omega_i$ denote its fixed weight, $c_{it}$ the supervision credit assigned to token $t$, and $M_i=\sum_t c_{it}$ its supervised mass. The implemented objective is
\begin{equation}
\mathcal L_{\rm credit}(\theta)=-\sum_i\frac{\omega_i}{M_i}\sum_t c_{it}\log \pi_\theta(a_{it}\mid x_{it}).
\label{eq:implemented-credit-training}
\end{equation}
Weights and normalization factors are frozen before optimization. The objective uses offline weighted supervision; it does not involve online policy optimization.

The maximum training window is 262,144 tokens. Seven trajectories require multiple windows, producing 2,404 windows from 2,397 unique trajectories. Windows preserve canonical turn and tool-batch boundaries. Repeated initial and left context receives zero loss, and all windows from a trajectory share its supervision denominator and optimizer batch. The retained trajectories contain 304,424,899 tokens before window preparation. The final training sequences contain 291,121,000 tokens and preserve all 6,177,605 supervised targets. These totals refer to different preparation stages; the main text reports the final training-sequence count.

\subsection{Training Configuration}
\label{supp:c-3}
Training uses full-parameter language-model fine-tuning with BF16 precision, Adam, a learning rate of $5\times10^{-6}$, and a global batch size of 16. One epoch completes in 150 optimizer steps. The distributed configuration uses 64 GPUs with tensor parallelism 4, context parallelism 8, and data parallelism 2. The final batch is padded with zero-loss entries to preserve the nominal batch normalization.

\subsection{Serialization and Target Selection}
\label{supp:c-4}
\label{app:serialization}
Each trajectory retains its source identifier, original messages and tool definitions, linked hypotheses, contribution evidence, and exposure metadata. A token mask records eligible original assistant outputs and zero-loss positions. Instructions and tool observations remain context; known failed actions, repeated context, and unavailable reasoning placeholders receive zero direct loss. Serialization preserves instruction order and excludes future information from earlier contexts. Unsupported instruction sequences are omitted from the training export while their source records remain available.

The data preparation groups shared memory, ancestry, resumed interactions, and recovered duplicates before assigning balancing coefficients. These groups are distinct from the scientific families used to cap hypothesis credit. Targets are counted once across windows, and each trajectory retains one denominator. The fixed token coefficient under uniform trajectory sampling is $N\omega_i/M_i$, where $N=2{,}397$ and $M_i$ is the trajectory's total supervised mass. Zero-loss padding preserves the nominal batch size at the end of the epoch.

The implemented token credit is constant at $c_{it}=0.5$ on eligible targets and zero elsewhere. This positive constant cancels between the numerator and $M_i$, giving the eligible-token mean in Eq.~\eqref{eq:sft}. Thus the main-text objective and Eq.~\eqref{eq:implemented-credit-training} describe the same normalization. The weights specify relative loss contributions, not guaranteed shares of gradient magnitude.

Preparation checks cover source integrity, token/mask alignment, unique target coverage, fixed trajectory denominators, and domain-weight accounting. They validate the training representation; scientific correctness is assessed separately. In particular, conserving supervised targets and credit does not establish that the underlying hypothesis is scientifically supported. The source records and private research artifacts are not distributed with this supplement.

\subsection{Windowing: Preserved Targets and Conditional Equivalence}
\label{supp:c-5}
\label{app:window-proof}
\begin{samepage}
\begin{proposition}[Objective preservation under windowing]
\label{prop:window-objective}
Let $U_i$ be the eligible target positions of trajectory $i$, with $|U_i|>0$, and let $\{U_{iw}\}_w$ partition $U_i$. If each target retains its original conditioning context, all windows use the same fixed coefficient $\omega_i$, and normalization uses $|U_i|$, then
\[
 -\sum_i\frac{\omega_i}{|U_i|}\sum_w\sum_{u\in U_{iw}}
 \log\pi_\theta(a_{iu}\mid x_{iu})
 =-\sum_i\frac{\omega_i}{|U_i|}\sum_{u\in U_i}
 \log\pi_\theta(a_{iu}\mid x_{iu}).
\]
The windowed and original objectives therefore have identical gradients wherever defined.
\end{proposition}
\end{samepage}

\paragraph{Proof and implementation scope.}
The target sets $U_{iw}$ partition $U_i$, so each target appears exactly once. With identical conditioning contexts, summing their log-likelihoods over windows gives the full-trajectory sum. Multiplication by the fixed coefficient $-\omega_i/|U_i|$ proves equality of the objectives and their gradients. Repeated context has zero loss. A constant positive token credit cancels in the numerator and denominator.

The conditioning assumption is stronger than target conservation. The implemented 256K export retains initial instructions and bounded overlap, so a window can omit earlier context from a longer trajectory. It preserves target counts and fixed normalization but does not generally preserve full-history likelihoods or gradients. The proposition establishes equivalence only when conditioning is unchanged; independently renormalizing each window would also violate its assumptions.

\section{DFM-Bench Evaluation Protocol}
\label{app:dfm-protocol}
DFM-Bench evaluates five objects: individual hypotheses, revisions, development paths, research runs, and scientific contributions. This appendix defines their scoring, evidence requirements, and coverage. The version of the rubric and evaluator is retained with each assessment; results from different applicability rules are not pooled without identifying those differences.

\begin{table}[htbp]
\small\centering
\caption{\textbf{The five DFM-Bench evaluation objects.} Scientific quality is evaluated at each level using evidence appropriate to that object. Delivery maturity is a separate axis.}
\label{tab:dfm}
\begin{tabularx}{\linewidth}{@{}p{2.3cm}p{2.1cm}YY@{}}
\toprule Level & Object & Core scientific question & Outputs \\\midrule
Hypothesis & Frozen candidate plus evidence & Is this a worthwhile, grounded, testable research opportunity? & Eight-dimension profile; Q /10; unknowns and review coverage. \\[2mm]
Revision & Actual parent--child pair & Did this step advance understanding, repair an important defect, or repeat prior work? & Substantive progress, necessary repair, repetition, insufficient evidence; numeric $\Delta Q$ separately. \\[2mm]
Development path & Ordered hypothesis path & Does inquiry accumulate evidence and revise decisions coherently? & Seven-axis L1--L7 profile; complete, scope-matched composite /100. \\[2mm]
Research run & Full research graph and decisions & Does the portfolio address meaningful opportunities and allocate effort well? & Independent T1--T7 profile; coverage, synthesis, allocation, and frontier diagnostics. \\[2mm]
Scientific contribution & Deduplicated contribution family & What question, distinction, evidence, negative result, or uncertainty resolution was added? & Contribution ledger, evidence maturity, novelty scope, and adjudication status. \\
\bottomrule
\end{tabularx}
\end{table}

\subsection{Hypothesis Quality and Automated Review}
\label{supp:d-1}
The hypothesis profile uses eight dimensions: importance, novelty, grounding, falsifiability, data resolvability, feasibility, alternatives, and actionability. Their weights are 20, 20, 15, 10, 10, 10, 10, and 5, respectively, on a 1--10 scale. These are frozen internal protocol conventions, not empirically established optimal weights.

A rubric judge, a domain-scientist role, and a methods-reviewer role assess the same evidence independently of one another's grades. The current panel uses separate prompts to the same model family. It therefore measures correlated AI perspectives, not three independent expert opinions. Each dimension takes the median when at least two reviewers provide a known score. A complete quality score $Q$ requires all three reviews, majority eligibility, and all eight dimension medians. The dimension weights sum to 100 and define a weighted mean on the 1--10 scale. Unknown evidence remains unknown rather than becoming zero.

Review packets use source-verbatim cards and traceable supporting material. They distinguish a proposal's scientific importance from its maturity: an important question need not already have a positive result. Conversely, a completed script does not demonstrate a consequential finding. The packet's available literature bounds novelty judgments; global novelty is not inferred from limited retrieval.

\subsection{Revisions, Development Paths, and Research Runs}
\label{supp:d-2}
Revision evaluation retains the four categorical judgments used by the progress contract. Substantive-progress rates, useful-step rates including repairs, repetition, and insufficient-evidence coverage are reported separately. Numeric $\Delta Q=Q_{\rm child}-Q_{\rm parent}$ is defined when both automated quality scores are available.

Development-path evaluation assesses explicitly ordered paths. The recorded protocol selects one longest-depth path to each terminal node, breaking ties by identifier, and excludes singleton paths. At crossovers, off-path parents remain context. Coverage therefore counts evaluated paths rather than every possible route through the graph. The seven weights are 5/30/20/20/10/10/5. Research-run evaluation independently inspects the entire graph using weights 25/15/15/20/10/10/5. Both use 0--4 dimension anchors, converted to /100 only under the required completeness and applicability rules.

A legitimately inapplicable dimension differs from an unknown one. Restricted-scope composites are grouped by applicability mask; missing required evidence prevents a full composite. Research-run scores are not averages of development-path scores. Shared ancestors also make path-level measurements correlated, precluding naive independent-sample interpretations.

\subsection{Scientific Contribution, Delivery, and Calibration}
\label{supp:d-3}
Scientific-contribution assessment groups related candidates into contribution families and records what was added beyond the starting evidence. Family membership does not imply confirmation; maturity cannot transfer automatically from one member to all others. Negative or inconclusive evidence can matter when it changes an important scientific decision. Human adjudication and independent evidence are needed for stronger claims of novelty and discovery.

Delivery is recorded alongside this scientific profile: proposed, selected, compiled, readiness-passed, executed, and independently reviewed. An unattempted check is distinct from a failure. The protocol does not combine delivery maturity and scientific scores into one overall number.

Calibration concerns the meaning and consistency of judgments. The implemented workflow freezes domain anchors, packets, aggregation, disagreements, and missingness. Same-packet comparisons against fresh-context AI references provide development diagnostics. Human calibration requires independent expert judgments, adjudication, and unseen acceptance cases; it remains pending. Prompt-role agreement alone does not establish validity.

The DFM framework's distinction between current knowledge progress and future capability growth \citep{dfm} applies across all five objects. A good final hypothesis does not establish a sound development path, and a disciplined research run does not establish that the trained model improves on future research runs. Frontier plots are useful diagnostics, but a retrospective maximum over final-snapshot scores is monotone by construction and cannot demonstrate learning.

\subsection{Development-Path and Research-Run Rubrics}
\label{supp:d-4}
Dimension scores use 0--4 evidence anchors: absent or contradicted, recognized with major gaps, partly substantiated, consistently substantiated, and comprehensive within the declared scope. Domain-specific evidence requirements qualify these anchors. The following labels summarize the clinical rubric; weights are fixed protocol choices rather than calibrated universal utilities.
\begin{center}
\small
\begin{tabularx}{\linewidth}{@{}lcY@{}}
\toprule Dimension & Weight & Criterion \\\midrule
L1 & 5\% & Relevant disciplinary connections \\
L2 & 30\% & Technical and knowledge exploration \\
L3 & 20\% & Cumulative scientific progress \\
L4 & 20\% & Evidential and inferential discipline \\
L5 & 10\% & Scientific continuity and synthesis \\
L6 & 10\% & Adaptive continuation and stopping \\
L7 & 5\% & Efficient use of the path's budget \\
\bottomrule
\end{tabularx}
\end{center}

\begin{center}
\small
\begin{tabularx}{\linewidth}{@{}lcY@{}}
\toprule Dimension & Weight & Criterion \\\midrule
T1 & 25\% & Disciplinary coverage and integration \\
T2 & 15\% & Technical and knowledge exploration \\
T3 & 15\% & Cumulative scientific progress \\
T4 & 20\% & Evidential and inferential discipline \\
T5 & 10\% & Portfolio synthesis \\
T6 & 10\% & Adaptive scientific decisions \\
T7 & 5\% & Budget-aware exploration \\
\bottomrule
\end{tabularx}
\end{center}

\subsection{Assessment Records and Missing Evidence}
\label{supp:d-5}
\label{app:assessment-records}
Each assessment identifies its object, source snapshot, rubric and protocol versions, supporting evidence, scores, rationale, and review status. Hypothesis assessments use a frozen proposal and its evidence. Revisions require actual parent--child pairs; development paths require ordered ancestry; research-run assessment requires the full graph and recorded decisions. Agent trajectories supply evidence about the work behind these objects.

Scores are recorded as known, unknown, or inapplicable under the rubric. Unknown values are not replaced by zero. An incomplete assessment is reported with its coverage and does not receive a complete composite. Resource-allocation judgments require recorded budgets and opportunity information; proposal counts do not reconstruct missing budgets. Contribution records additionally identify the scientific question, the increment beyond starting evidence, and its status as supported, contradicted, inconclusive, or unresolved. Evidence maturity distinguishes proposed, source-grounded, diagnostically examined, study-completed, and independently checked contributions.

\subsection{Benchmark Coverage Definitions}
\label{supp:d-6}
\label{app:dfm-comparison}
\Cref{tab:benchmark-coverage} compares explicit evaluation constructs, not the capabilities of agents that happen to run on a benchmark. Hypothesis merit concerns scientific value and testability; contribution maturity distinguishes proposed, examined, and independently supported contributions. Evidence use concerns grounding analyses or claims in available evidence. Revision assessment compares a change against a predecessor; development-path assessment examines cumulative inquiry; whole-run allocation assesses decisions among research directions. A cross means no explicit assessment under these definitions, not that the benchmark lacks scientific value. Partial coverage identifies a related measure with narrower scope.

AgentIdeaBench scores final ideas with literature-verified critics; its grounding assessment is partial evidence-use coverage. Its generation-time refinement experiments do not constitute an explicit revision or ancestry-path evaluation object \citep{agentideabench}. DiscoveryBench evaluates reference-aligned hypotheses and analytical facets for supplied discovery goals; this provides partial hypothesis-merit and evidence-use coverage, rather than an open-ended merit or contribution-maturity rubric \citep{discoverybench}. BLADE compares conceptual variables, transformations, and statistical models against expert analyses, providing partial evidence-use coverage through analytical decisions \citep{blade}. ScienceAgentBench assesses generated programs, outputs, and task-stage rubrics; this is partial evidence-use coverage rather than evaluation of evolving scientific claims \citep{scienceagentbench}.

OEB explicitly assesses evidence, experiments, and responses to contradicted ideas. Its inquiry grouping and search-topology measures supply partial development-path coverage; its budget and concurrency profiles supply partial whole-run allocation coverage because they describe research style without ranking one style above another. It does not supply the proposal-merit or contribution-maturity assessments defined here \citep{oeb}. DFM-Bench's checks indicate implemented rubric coverage conditional on sufficient records, not independent expert validation or empirical superiority over these instruments.

\section{Human Assessment Protocol}
\label{app:human-assessment}
Human assessment examines the value and design quality of proposed investigations, with automated scores hidden. Overall value, criterion ratings, recommendations, and design-flaw verdicts are separate judgments.

\subsection{Review Cohort and Presentation}
\label{supp:e-1}
\label{app:human-cohort}
The review involved 27 medical doctors, medical students, and bioscience PhD students with biomedical or clinical research backgrounds, under three senior supervisors. The fixed hypothesis-atlas cohort contains 292 proposals: HCC (52), MIMIC-IV (60), eICU (60), UK Biobank (55), GEO therapeutic targets (35), and GEO disease mechanisms (30). The groups span the four evaluated life-science domains.

English proposal packets included the scientific question, investigation plan, data requirements, alternative explanations, and citations. Model identity, execution provenance, and automated grades were hidden. Proposals were presented in randomized order, and reviewers assessed each packet without consulting its revision history. Figure~\ref{fig:expert-review} illustrates the interface.

\subsection{Rating Scales and Recommendations}
\label{supp:e-2}
\label{app:human-scales}
Reviewers first recorded overall research value on a five-point scale from not worth pursuing to top priority, together with a recommendation to prioritize, revise, defer, or reject. They then rated six criteria on anchored four-point scales: importance, advance over known work, plausible and evidence-bounded claims, data fit, alternative explanations, and informative results. A separate verdict classified consequential design flaws as absent, fixable, or fatal. Reviewers could select ``Can't judge'' when evidence or expertise was insufficient; flaw checklists and explanatory comments were optional.

The five-point overall judgment is distinct from the four-point criterion profile. Recommendations to revise or defer do not imply endorsement, and design-flaw verdicts concern the investigation plan.

\subsection{Revision Outcomes and Interpretation}
\label{supp:e-3}
\label{app:human-revisions}
The revision analysis compares human-rated quality for a revised hypothesis and its actual predecessor. Positive, zero, and negative score changes define improved, unchanged, and worsened revisions, respectively. The reported mean change and outcome proportions characterize the assessed parent--child pairs. They do not estimate the effect of a model update across independent investigations.

Figure~\ref{fig:human-quality-progress} separates criterion profiles and revision outcomes. The MIMIC-IV case (Figure~\ref{fig:human-revision-case}) shows six-criterion means along a selected path; it does not imply that all branches improve. Shared ancestry makes revision observations dependent. This assessment is distinct from calibration of the automated evaluator against independently adjudicated, unseen cases.

\section{Literature-Inspiration Evaluation}
\label{app:reference-actionability}
This appendix defines the retrospective literature comparison used in RQ4. ``Inspiration'' denotes a concrete scientific connection: a reference after the recorded cutoff provides an experimental design, comparator, measurement, or testable evidence gap relevant to developing or testing a generated hypothesis. It does not imply that the publication's authors encountered the hypothesis or were causally influenced by it.

\subsection{Reference Corpus and Temporal Boundaries}
\label{supp:f-1}
\label{app:inspiration-corpus}
One model-blind reference corpus was constructed from deterministic source-only anchors. A completed citation wave followed 420 backward and forward citation traversals, yielding 17,790 citation edges and 16,440 resolved works. The same corpus was available to all model groups within the publication window from 1 April 2025 through 14 October 2026.

References were filtered by each group's recorded operational cutoff. Exact-day boundaries were retained for GPT serving identifiers. Month-level boundaries were used for Qwen (April 2025, a provisional family proxy), Claude Opus 5 (May 2026), DeepSeek V4.1 (September 2026, a release-month proxy), and GLM-5.3 (August 2026, a release-month proxy). These boundaries make the comparison reproducible; they do not identify internal model knowledge cutoffs. 

\subsection{Retrieval and Automated Judgment}
\label{supp:f-2}
\label{app:inspiration-audit}
Candidate retrieval required the best reference card to share at least three non-stopword terms with the hypothesis. This lexical gate determines retrieval eligibility, not semantic relevance. GPT-6 Luna supplied the semantic judgments, with 145 successful judgments from 149 sampled records. Four records lacked successful judgments. Exact duplicate hypothesis--reference inputs shared a judgment, which was propagated to 903 registered hypothesis versions. The resulting labels are correlated automated assessments rather than independent expert reviews.

The inspiration rate uses all generated hypothesis versions as its denominator, as defined in Eq.~\eqref{eq:reference-actionability-rate}. Only versions receiving a positive judgment contribute to the numerator. Consequently, absence of a positive label may reflect retrieval or audit coverage as well as lack of a qualifying connection. The underlying audit calls this endpoint reference actionability; the report uses the operational term inspiration rate consistently.

\subsection{Population Accounting}
\label{supp:f-3}
\label{app:inspiration-population}
DiscoveryRSI-27B pools two production runs. The overall numerator and denominator are 74/1,176; domain counts are shown below. The comparison groups are DeepSeek (14/257), GLM (1/36), GPT (59/2,341), and Claude (3/131). Qwen original remains in the underlying audit but is omitted from the main presentation figure.

\begin{table}[!htbp]
\centering\small
\caption{\textbf{DiscoveryRSI-27B literature-inspiration counts.} Numerators count hypothesis versions with positive judgments; denominators include all generated versions.}
\label{tab:inspiration-counts}
\begin{tabular}{@{}lrrr@{}}
\toprule
Domain & Positive & Generated & Rate \\
\midrule
Clinical and population health & 13 & 557 & 2.3\% \\
Therapeutic target prioritization & 30 & 178 & 16.9\% \\
Disease mechanisms & 17 & 251 & 6.8\% \\
Population multi-omics & 14 & 190 & 7.4\% \\
\midrule
Total & 74 & 1,176 & 6.3\% \\
\bottomrule
\end{tabular}
\end{table}

Group sizes and domain composition differ, and some strata are small. Versions from the same research family and reused judgments are dependent. The report therefore presents descriptive percentages without significance tests.

\subsection{Claude Opus 5: Retrieval Sensitivity}
\label{supp:f-4}
\label{app:claude-retrieval}
The targeted analysis in RQ4 separates retrieval coverage from positive judgments conditional on retrieval. Only 15/131 Claude proposals passed the retrieval gate (11.5\%), compared with 355/1,176 DiscoveryRSI proposals (30.2\%). Among retrieved proposals, the positive fractions were similar: 20.0\% and 20.8\%, respectively.

Claude's median proposal length was 41,134 characters, whereas retrieval used the first 6,000 characters and semantic review used the first 2,500. Provenance and revision notes could therefore displace scientific content. A diagnostic substituted proposal titles as queries while preserving the corpus, scoring method, date rule, and overlap threshold. Claude's retrieval count increased from 15 to 95. This establishes sensitivity to representation, not additional positive semantic matches. A structured scientific representation and renewed judgments across all groups would be needed to assess the effect on the final comparison.

The audit observes literature correspondence, not researcher exposure, adoption, study initiation, or causal influence. Prospective evaluation would freeze generated hypotheses before researcher review and follow subsequent investigation plans, experiments, and publications.

\end{document}